\documentclass[12pt]{article}
\usepackage{graphicx,amsmath,amsfonts,amssymb,natbib,color}
\usepackage{multirow}
\usepackage{algorithm}
\usepackage{algorithmic}
\usepackage{geometry}
\title{\LARGE Variational Particle Approximations}

\author{
       \large  Ardavan Saeedi\footnote{First two authors contributed equally.}  \\
       \small \texttt{ardavans@mit.edu} \\
      \small   CSAIL \\
       \small   Massachusetts Institute of Technology\\
       \and
       \large  Tejas D. Kulkarni\small* \\
       \small  \texttt{tejask@mit.edu} \\
       \small   Department of Brain \& Cognitive Sciences\\
      \small  Massachusetts Institute of Technology\\
       \vspace{0.5mm}
       \and
       \large  Vikash K. Mansinghka \\
      \small   \texttt{vkm@mit.edu} \\
       \small  Department of Brain \& Cognitive Sciences\\
       \small  Massachusetts Institute of Technology\\
       \vspace{0.5mm}
       \and
       \large Samuel J. Gershman \\
       \small  \texttt{gershman@fas.harvard.edu} \\
      \small  {Department of Psychology and Center for Brain Science}\\
       \small  Harvard University
       }

\begin{document}
\date{}
\maketitle

\begin{abstract}
Approximate inference in high-dimensional, discrete probabilistic models is a central problem in computational statistics and machine learning. This paper describes discrete particle variational inference (DPVI), a new approach that combines key strengths of Monte Carlo, variational and search-based techniques. DPVI is based on a novel family of particle-based variational approximations that can be fit using simple, fast, deterministic search techniques. Like Monte Carlo, DPVI can handle multiple modes, and yields exact results in a well-defined limit. Like unstructured mean-field, DPVI is based on optimizing a lower bound on the partition function; when this quantity is not of intrinsic interest, it facilitates convergence assessment and debugging. Like both Monte Carlo and combinatorial search, DPVI can take advantage of factorization, sequential structure, and custom search operators. This paper defines DPVI particle-based approximation family and partition function lower bounds, along with the sequential DPVI and local DPVI algorithm templates for optimizing them. DPVI is illustrated and evaluated via experiments on lattice Markov Random Fields, nonparametric Bayesian mixtures and block-models, and parametric as well as non-parametric hidden Markov models. Results include applications to real-world spike-sorting and relational modeling problems, and show that DPVI can offer appealing time/accuracy trade-offs as compared to multiple alternatives.
\end{abstract}


\section{Introduction}

Monte Carlo methods are based on the idea that one can approximate a complex distribution with a set of stochastically sampled particles. The flexibility and variety of Monte Carlo methods have made them the workhorse of statistical computation \citep{robert04}. However, their success relies critically on having available a good sampler, and designing such a sampler is often challenging.

In this paper, we rethink particle approximations over discrete hypothesis spaces from a different perspective. Suppose we got to pick where to place the particles in the hypothesis space; where would we put them? Intuitively, we would want to distribute them in such a way that they cover high probability regions of the target distribution, but without the particles all devolving onto the mode of the distribution. This problem can be formulated precisely within the framework of variational inference \citep{wainwright08}, which treats probabilistic inference as an optimization problem over a set of distributions. We derive a coordinate ascent update for particle approximations that iteratively minimizes the Kullback-Leibler (KL) divergence between the particle approximation and the target distribution.

After introducing our general framework, we describe how it can be applied to filtering and smoothing problems. We then show experimentally that variational particle approximations can overcome a number of problems that are challenging for conventional Monte Carlo methods. In particular, our approach is able to produce a diverse, high probability set of particles in situations where Monte Carlo and mean-field variational methods sometimes degenerate.

\section{Background}

Consider the problem of approximating a probability distribution $P(x)$ over discrete latent variables $x = \{x_1,\ldots,x_N\}, x_n \in \{1,\ldots,M^k_n\}$, where the target distribution is known only up to a normalizing constant $Z$: $P(x) = f(x)/Z$. We will refer to $f(x) \geq 0$ as the \emph{score} of $x$ and $Z$ as the \emph{partition function}. We further assume that $P(x)$ is a Markov network defined on a graph $G$, so that $f(x)$ factorizes according to:
\begin{align}
f(x) = \prod_{c} f_c(x_c),
\end{align}
where $c \subseteq \{1,\ldots,N\}$ indexes the maximal cliques of $G$.

\subsection{Importance sampling and sequential Monte Carlo}

A general way to approximate $P(x)$ is with a weighted collection of $K$ particles, $\{ x^1,\ldots,x^K \}$:
\begin{align}
P(x)  \approx Q(x)  = \sum_{k=1}^K w^k \delta[x,x^k],
\label{eq:particle}
\end{align}
where $x^k = \{x^k_1,\ldots,x^k_N\}, x^k_n \in \{1,\ldots,M^k_n\}$ and $\delta[\cdot,\cdot]=1$ if its arguments are equal and 0 otherwise. Importance sampling is a Monte Carlo method that stochastically generates particles from a proposal distribution, $x^k \sim \phi(\cdot)$, and computes the weight according to $w^k \propto f(x^k)/\phi(x^k)$. Importance sampling has the property that the particle approximation converges to the target distribution as $K \rightarrow \infty$ \citep{robert04}.

Sequential Monte Carlo methods such as particle filtering \citep{doucet01} apply importance sampling to stochastic dynamical systems (where $n$ indexes time) by sequentially sampling the latent variables at each time point using a proposal distribution $\phi(x_n|x_{n-1})$. This procedure can produce conditionally low probability particles; therefore, most algorithms include a resampling step which replicates high probability particles and kills off low probability particles. The downside of resampling is that it can produce degeneracy: the particles become concentrated on a small number of hypotheses, and consequently the effective number of particles is low.

\subsection{Variational inference}

Variational methods \citep{wainwright08} define a parametrized family of probability distributions $\mathcal{Q}$ and then choose $Q \in \mathcal{Q}$ that maximizes the \emph{negative variational free energy}:
\begin{align}
\mathcal{L}[Q] = \sum_x Q(x) \log \frac{f(x)}{Q(x)}.
\end{align}
The negative variational free energy is related to the partition function $Z$ and the KL divergence through the following identity:
\begin{align}
\log Z = \mbox{KL}[Q || P] + \mathcal{L}[Q],
\end{align}
where
\begin{align}
\mbox{KL}[Q || P] = \sum_x Q(x) \log \frac{Q(x)}{P(x)}.
\end{align}
Since $\mbox{KL}[Q || P] \geq 0$, the negative variational free energy is a lower bound on the log partition function, achieving equality when the KL divergence is minimized to 0. Maximizing $\mathcal{L}[Q]$ with respect to $Q$ is thus equivalent to minimizing the KL divergence between $Q$ and $P$.

Unlike the Monte Carlo methods described in the previous section, variational methods do not in general converge to the target distribution, since typically $P$ is not in $\mathcal{Q}$. The advantage of variational methods is that they guarantee an improved bound after each iteration, and convergence is easy to monitor (unlike most Monte Carlo methods). In practice, variational methods are also often more computationally efficient.

We next consider particle approximations from the perspective of variational inference. We then turn to the application of particle approximations to inference in stochastic dynamical systems.

\section{Variational particle approximations}

Variational inference can be connected to Monte Carlo methods by viewing the particles as a set of variational parameters parameterizing $Q$. For the particle approximation defined in Eq. \ref{eq:particle}, the negative variational free energy takes the following form:
\begin{align}
\mathcal{L}[Q] = \sum_{k=1}^K w^k \log \frac{f(x^k)}{w^k V^k},
\end{align}
where $V^k = \sum_{j=1}^K \delta[x^j,x^k]$ is the number of times an identical replica of $x^k$ appears in the particle set. We wish to find the set of $K$ particles and their associated weights that maximize $\mathcal{L}[Q]$, subject to the constraint that $\sum_{k=1}^K w^k=1$. This constraint can be implemented by defining a new functional with Lagrange multiplier $\lambda$:
\begin{align}
\tilde{\mathcal{L}}[Q] = \mathcal{L}[Q] + \lambda \left (\sum_{k=1}^K w^k - 1\right).
\end{align}
Taking the functional derivative of the Lagrangian with respect to $w^k$ and equating to zero, we obtain:
\begin{align}
\frac{\partial \tilde{\mathcal{L}}[Q] }{\partial w^k} &= \log f(x^k) - \log w^k - \log V^k + \lambda - 1 = 0 \nonumber \\
&\Longrightarrow w^k = Z_Q^{-1} f(x^k)/V^k,
\end{align}
where
\begin{align}
Z_Q =  \exp(\lambda-1)^{-1} = \sum_{k=1}^K \frac{f(x^k)}{V^k}.
\end{align}
We can plug the above result back into the definition of $\mathcal{L}[Q]$:
\begin{align}
\mathcal{L}[Q] 
&= Z^{-1}_Q \sum_{k=1}^K \frac{f(x^k)}{V^k} \log \frac{f(x^k) V^k}{Z^{-1}_Q f(x^k) V^k} \nonumber \\
&= Z^{-1}_Q \sum_{k=1}^K \frac{f(x^k)}{V^k}  \log Z_Q \nonumber \\
&= \log Z_Q
\label{eq:elbo}
\end{align}
Thus, $\mathcal{L}[Q]$ is maximized by choosing the $K$ values of $x$ with the highest score. The following theorem shows that allowing $V^k >1$ (i.e., having replica particles) can never improve the bound.

\vspace{1mm}
\noindent \textbf{Theorem:}
\textit{Let $Q$ and $Q'$ denote two particle approximations, where $Q$ consists of unique particles ($V^k=1$ for all $k$) and $Q'$ is identical to $Q$ except that particle $x^j$ is replicated $V^j$ times (displacing $V^j$ other particles with cumulative score $F$). For any choice of particles, $\mathcal{L}[Q] \geq \mathcal{L}[Q']$.}
\vspace{1mm}

\noindent \textbf{Proof:}
We first apply Jensen's inequality to obtain an upper bound on $\mathcal{L}[Q']$:
\begin{align}
\mathcal{L}[Q'] \leq \log \sum_{k=1}^K w^k Z_Q = \log \sum_{k=1}^K \frac{f(x^k)}{V^k}.
\end{align}
Since $\mathcal{L}[Q] = \log Z_Q$, we wish to show that $Z_Q \geq \sum_{k=1}^K \frac{f(x^k)}{V^k}$. All the particles in $Q$ and $Q'$ are identical except for $x^j$ and the $V^j$ particles in $Q$ that were displaced by replicas of $x^j$ in $Q'$; thus we only need to establish that $f(x^j) + F \geq \frac{V^j f(x^j)}{V^j} = f(x^j)$. Since the score can never be negative, $F \geq 0$ and the inequality holds for any choice of particles. $\blacksquare$
\vspace{1mm}
\begin{algorithm}
\caption{Discrete particle variational inference}
\begin{algorithmic}[1]
\STATE{/*$N$ is the number of latent variables */}
\STATE{/*$x^k$ is the set of all latent variables for the $k$th particle:  $x^k = \{x^k_1,\ldots,x^k_N\} $ */}
\STATE{/*$M^k_n$ is the support of latent variable $x^k_n$ */}
\STATE{\textbf{Input}: initial particle approximation $Q$ with $K$ particles, tolerance $\epsilon$}
\WHILE{$|\mathcal{L}[Q] - \mathcal{L}[Q']|>\epsilon$}
\FOR{$n=1$ \TO $N$}
\STATE{$\mathcal{X} = \emptyset$}
\FOR{$k=1$ \TO $K$}
\STATE{Copy particle $k$: $\tilde{x}^k \leftarrow x^k$}
\FOR{$m=1$ \TO $M^k_n$}
\STATE{Modify particle: $\tilde{x}_n^k \leftarrow m$}
\STATE{Score $\tilde{x}^k$ using Eq. \ref{eq:score}}
\STATE{$\mathcal{X} \leftarrow \mathcal{X} \cup (\tilde{x}^k, f(\tilde{x}^k))$}
\ENDFOR
\ENDFOR
\STATE{Select the $K$ particles from $\mathcal{X}$ with the largest scores}
\STATE{Construct new particle approximation $Q'(x) = \sum_{k=1}^K w^k \delta[x,x^k]$}
\STATE{Compute variational bound $\mathcal{L}[Q']$ using Eq. \ref{eq:elbo}}
\ENDFOR
\ENDWHILE
\RETURN particle approximation $Q'$
\end{algorithmic}
\end{algorithm}

The variational bound can be optimized by coordinate ascent, as specified in Algorithm 1, which we refer to as \emph{discrete particle variational inference} (DPVI). This algorithm takes advantage of the fact that when optimizing the bound with respect to a single variable, only potentials local to that variable need to be computed. In particular, let $\tilde{x}^k$ be a replica of $x^k$ with a single-variable modification, $\tilde{x}_n^k=m$. We can compute the unnormalized probability of this particle efficiently using the following equation:
\begin{align}
f(\tilde{x}^k) = f(x^k) \frac{\mathcal{F}_n(\tilde{x}^k)}{\mathcal{F}_n(x^k)}
\label{eq:score}
\end{align}
where $\mathcal{F}_n(x) = \prod_{c: n \in c} f_c(x_c)$. The variational bound for the modified particle can then be computed using Eq. \ref{eq:elbo}. Particles can be initialized arbitrarily. When repeatedly iterated, DPVI will converge to a local maximum of the negative variational free energy. Note that in principle more sophisticated methods can be used to find the top $K$ modes \citep[e.g.,][]{flerova12,yanover03}; however, we have found that this coordinate ascent algorithm is fast, easy to implement, and very effective in practice (as our experiments below demonstrate).

An important aspect of this framework is that it maintains one of the same asymptotic guarantees as importance sampling: $Q$ converges to $P$ as $K \rightarrow \infty$, since in this limit DPVI is equivalent to exact inference. Thus, DPVI combines advantages of variational methods (monotonically decreasing KL divergence between $Q$ and $P$) with the asymptotic correctness of Monte Carlo methods. The asymptotic complexity of DPVI in the sequential setting is $O(SNK)$ where $S$ is the maximum support size of the latent variables. For the iterative update of the particles the complexity is $O(TCSK)$, where $T$ is the maximum number of iterations until convergence and $C$ is the maximum clique size. In our experiments, we empirically observed that we only need a small number of iterations and particles in order to outperform our baselines. 

\section{Filtering and smoothing in hidden Markov models}

We now describe how variational particle approximations can be applied to filtering and smoothing in hidden Markov models (HMMs). Consider a hidden Markov model with observations $y = \{ y_1,\ldots,y_N \}$ generated by the following stochastic process: 
\begin{align}
P(y,x,\theta) = P(\theta) \prod_{n} P(y_n|x_n,\theta) P(x_n|x_{n-1},\theta),
\end{align}
where $\theta$ is a set of transition and emission parameters. We are particularly interested in \emph{marginalized} HMMs where the parameters are integrated out: $P(y,x) = \int_{\theta} P(y,x,\theta) d\theta$. This induces dependencies between observation $n$ and all previous observations, making inference challenging.

Filtering is the problem of computing the posterior over the latent variables at time $n$ given the history $y_{1:n}$. To construct the variational particle approximation of the filtering distribution, we need to compute the product of potentials for variable $n$:
\begin{align}
\mathcal{F}_n(x) = P(y_n|x_{1:n},y_{1:n-1})P(x_n|x_{1:n-1}).
\label{eq:filter}
\end{align}
We can then apply the coordinate ascent update described in the previous section. This update is simplified in the filtering context due to the underlying Markov structure:
\begin{align}
f(\tilde{x}^k) = &f(x^k) P(y_n|x_n^k=m,x_{1:n-1},y_{1:n-1}) P(x^k_n=m|x_{1:n-1}).
\end{align}
At each time step, the algorithm selects the $K$ continuations (new variable assignments of the current particle set) that maximize the negative variational free energy.

Smoothing is the problem of computing the posterior over the latent variables at time $n$ given data from both the past and the future, $y_{1:N}$. The product of potentials is given by:
\begin{align}
\mathcal{F}_n(x) = P(y_n|x_{1:n},y_{-n})P(x_n|x_{-n}),
\label{eq:smooth}
\end{align}
where $x_{-n}$ refers to all the latent variables except $x_n$ (and likewise for $y_{-n}$). This potential can be plugged into the updates described in the previous section.

\begin{figure}
\centering
\begin{tabular}{cc}
\hspace{-5mm}\includegraphics[trim = 3mm -5mm 0mm 0mm, clip, scale = 0.5]{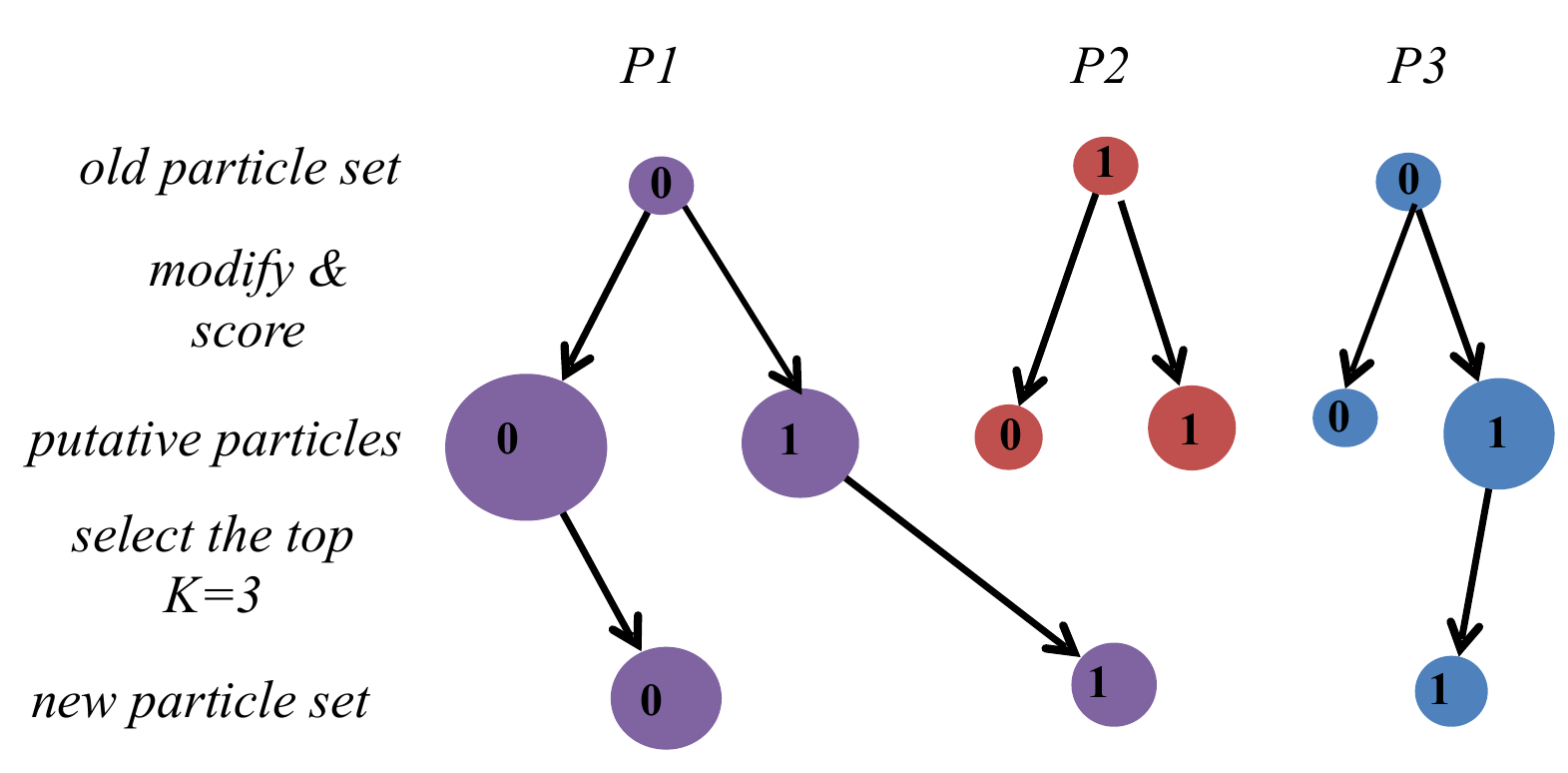} & \hspace{-13mm}
\includegraphics[trim = -30mm 0mm 0mm 0mm, clip, scale = 0.65]{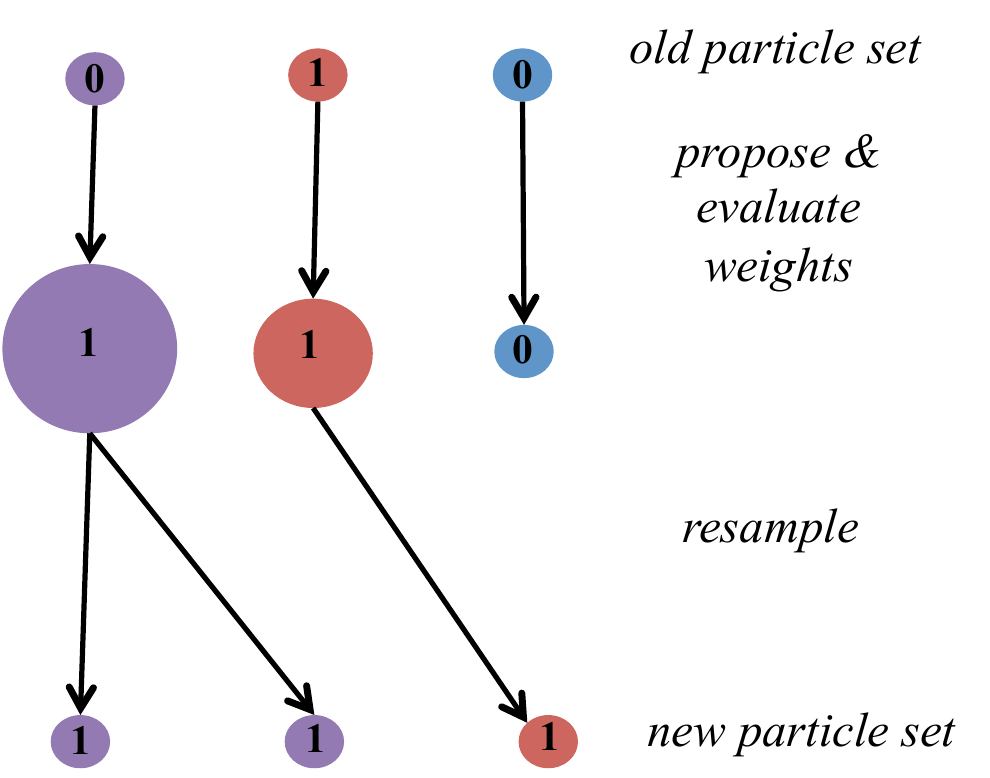}\\
\hspace{15mm} (A) DPVI  &\hspace{-17mm} (B) Particle Filtering \\
\end{tabular}
\caption{\textbf{Schematic of DPVI versus particle filtering for filtering problems}. Illustration of different filtering scenarios over 2 time steps in a binary state space with $K=3$ particles. The number in each circle indicates the binary value of the corresponding variable. Arrows indicate the evolution of the particles. (A) DPVI: The size of the putative particles represents the score of the particle. The $K$ continuations with highest score are selected for propagation to the next time step. The size of the new particle set corresponds to the normalized score. Particle $P1$ is split, $P2$ is deleted
and one putative particle from $P3$ is selected.  (B) Particle filtering: The size of the node represents the weight of the particle for the resampling step.}
\label{fig:schematic}
\end{figure}

To understand DPVI applied to filtering problems, it is helpful to contemplate three possible fates for a particle at time $n$ (illustrated in Figure \ref{fig:schematic}):
\begin{itemize}
\item \textbf{Selection}: A single continuation of particle $k$ has non-zero weight. This can be seen as a deterministic version of particle filtering, where the sampling operation is replaced with a max operation.
\item \textbf{Splitting}: Multiple continuations of particle $k$ have non-zero weight. In this case, the particle is split into multiple particles at the next iteration.
\item \textbf{Deletion}: No continuations of particle $k$ have non-zero weight. In this case, the particle is deleted from the particle set.
\end{itemize}
Similar to particle filtering with resampling, DPVI deletes and propagates particles based on their probability. However, as we show later, DPVI is able to escape some of the problems associated with resampling.

\section{Related work}

DPVI is related to several other ideas in the statistics literature:
\begin{itemize}
\item DPVI is a special case of a \emph{mixture mean-field variational approximation} \citep{jaakkola98,lawrence00}:
\begin{align}
Q(x) = \sum_{k=1}^K Q(k) \prod_{n=1}^N Q(x_n|k).
\end{align}
In DPVI, $Q(k) = w^k$ and $Q(x_n|k) = \delta[x_n,x_n^k]$. A distinct advantage of DPVI is that the variational updates do not require the additional lower bound used in general mixture mean-field, due to the intractability of the mean-field updates.
\item When $K=1$, DPVI is equivalent to \emph{iterated conditional modes} \citep[ICM;][]{besag86}, which iteratively maximizes each latent variable conditional on the rest of the variables.
\item DPVI is conceptually similar to nonparametric variational inference \citep{gershman12}, which approximates the posterior over a continuous state space using a set of particles convolved with a Gaussian kernel.
\item \citet{frank09} used particle approximations within a variational message passing algorithm. The resulting approximation is ``local'' in the sense that the particles are used to approximate messages passed between nodes in a factor graph, in contrast to the ``global'' approximation produced by DPVI, which attempts to capture the distribution over the entire set of variables.
\item \citet{ionides08} described a truncated version of importance sampling in which weights falling below some threshold are set to the threshold value. This is similar (though not equivalent) to the DPVI setting where latent variables are sampled exhaustively and without replacement.
\item Finally, DPVI is closely related to the problem of finding the $K$ most probable latent variable assignments \citep{flerova12,yanover03}. We view this problem through the lens of particle approximations, connecting it to both Monte Carlo and variational methods.
\end{itemize}

\section{Experiments}
\label{sec:experiment}

In this section, we compare the performance of DPVI to several widely used approximate inference algorithms, including particle filtering and variational methods. We first present a didactic example to illustrate how DPVI can sometimes succeed where particle filtering fails. We then apply DPVI to four probabilistic models: the Dirichlet process mixture model \citep[DPMM;][]{antoniak74,escobar95}, the infinite HMM \citep[iHMM;][]{beal02,teh06}, the infinite relational model \citep[IRM;][]{kemp2006learning} and the Ising model. 

\subsection{Didactic example: binary HMM}

\begin{figure}[ht]
\centering
\begin{tabular}{c}
(A)
\includegraphics[trim = 0mm -5mm 0mm 0mm, clip, scale = 0.7]{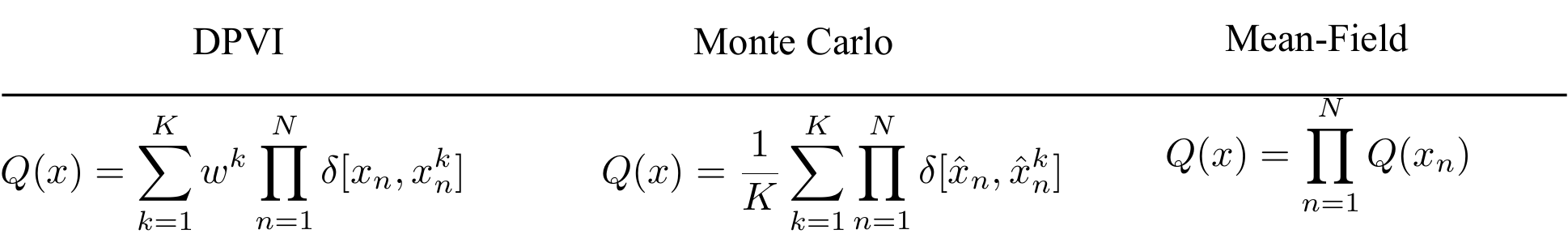}\\ 
(B)
\includegraphics[trim = 0mm 0mm 0mm 0mm, clip, scale = 0.7]{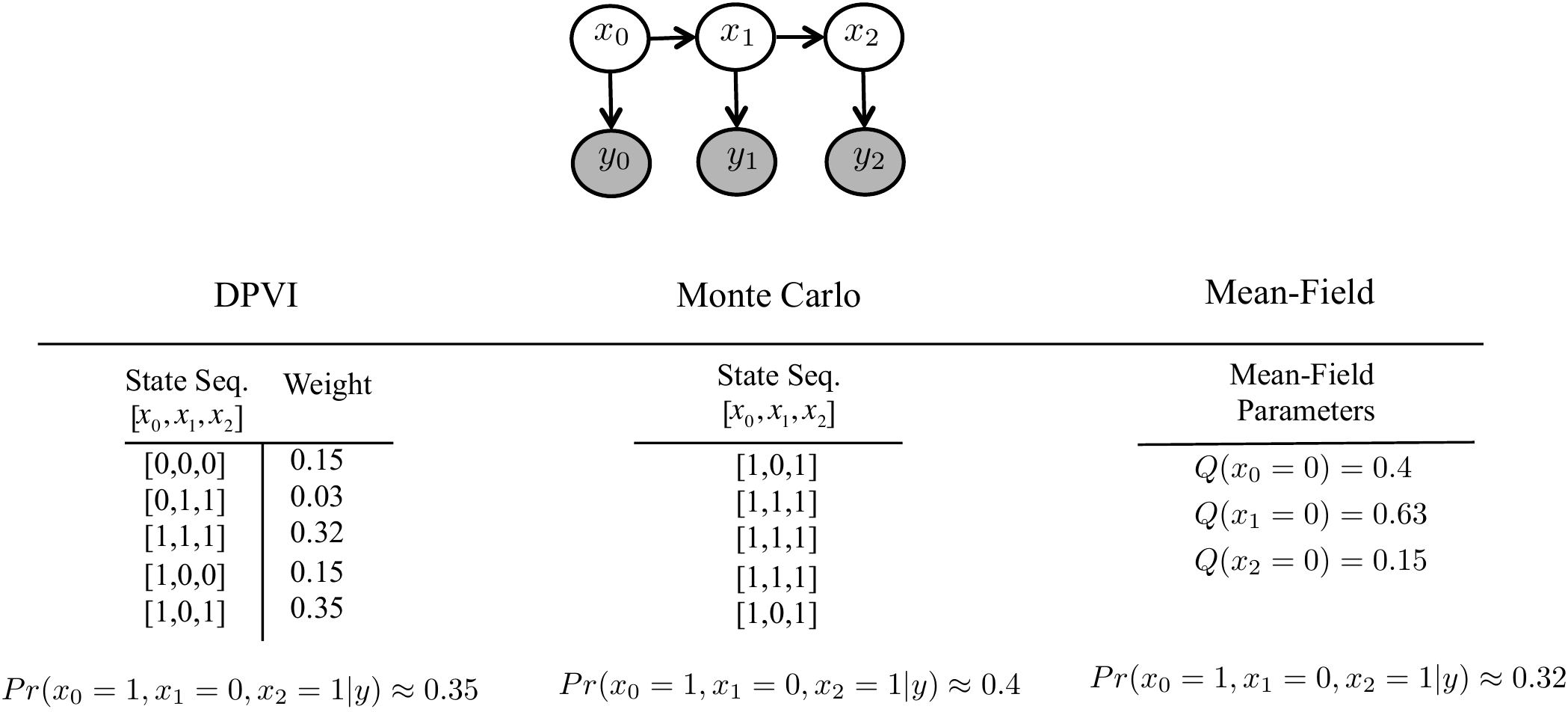}\\
\end{tabular}
\caption{\textbf{Comparison of approximate inference schemes}. (\emph{A}) Approximating families for DPVI, Monte Carlo and mean-field. (\emph{B}) Illustration of the differences between schemes in (A) on a binary HMM. } 
\label{fig:approx_family}
\end{figure}

As a didactic example, we use a simple HMM with binary hidden states ($x$) and observations ($y$):
\begin{align}
&P(x_{n+1}= 0|x_n = 0) = \alpha_{0} \nonumber \\
&P(x_{n+1}= 1|x_n = 1) = \alpha_{1} \nonumber \\
&P(y_n=0|x_n=0) = \beta_{0} \nonumber \\
&P(y_n=1|x_n=1) = \beta_{1},
\end{align}
with $\alpha_{0}$, $\alpha_{1}$, $\beta_{0}$, and $\beta_{1}$ all less than 0.5. We will use this model to illustrate how DPVI differs from particle filtering. Figure \ref{fig:approx_family} compares several inference schemes for this model.

\begin{figure}
\centering
\includegraphics[scale=0.45]{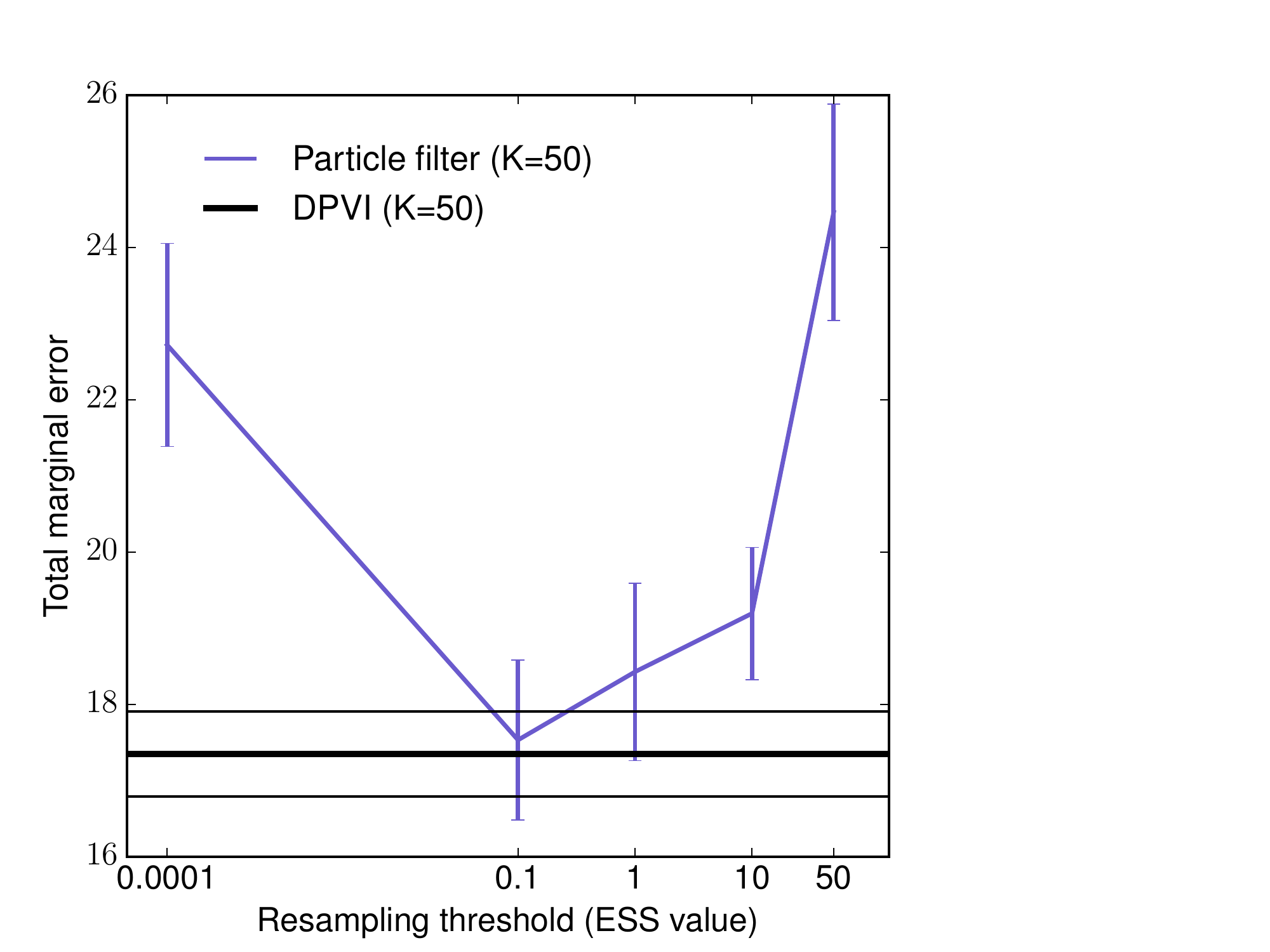}
\caption{\textbf{HMM with binary hidden states and observations.} Total marginal error computed for a sequence of length 200. For particle filtering the total error for every ESS value is averaged over 5 sequences generated from the HMM; in addition, for each sequence we reran the particle filter 5 times (thus 25 runs total). Note the logarithmic scale of the x-axis. Error bars and the thin black lines correspond to standard error of the mean.}
\label{fig:didactic_hmm}
\end{figure}

For illustration, we use the following parameters: $\alpha_{0} = 0.2$, $\alpha_{1} = 0.1$, $\beta_{0} = 0.3$, and $\beta_{1} = 0.2$. Suppose you observe a sequence generated from this model. For a sufficiently long sequence, a particle filter with resampling will eventually delete all conditionally unlikely particles, and thus suffer from degeneracy. On the other hand, without resampling the approximation will degrade over time because conditionally unlikely particles are never replaced by better particles. For this reason, it is sometimes suggested that resampling only be performed when the effective sample size (ESS) falls below some threshold. The ESS is calculated as follows:
\begin{align}
\mbox{ESS} =\frac{1}{\sum_{k=1}^K (w^k)^2}.
\end{align}
A low ESS means that most of the weight is being placed on a small number of particles, and hence the approximation may be degenerate (although in some cases this may mean that the target distribution is peaky). We evaluated particle filtering with multinomial resampling on synthetic data generated from the HMM described above. Approximation accuracy was measured by using the forward-backward algorithm to compute the hidden state posterior marginals exactly and then comparing these marginals to the particle approximation. Figure \ref{fig:didactic_hmm} shows performance as a function of ESS threshold, demonstrating that there is a fairly narrow range of thresholds for which performance is good. Thus in practice, successful applications of particle filtering may require computationally expensive tuning of this threshold.

In contrast, DPVI achieves performance comparable to the optimal particle filter, but without a tunable threshold. This occurs because DPVI uses an implicit threshold that is automatically tuned to the problem. Instead of resampling particles, DPVI deletes or propagates particles deterministically based on their relative contribution to the variational bound.

\subsection{Dirichlet process mixture model}

A DPMM generates data from the following process \citep{antoniak74,escobar95}:
\begin{align}
G \sim \mbox{DP}(\alpha,G_0), \quad \quad \theta_n|G \sim G, \quad \quad y_n|\theta_n \sim F(\theta_n), \nonumber
\end{align}
where $\alpha \geq 0$ is a concentration parameter and $G_0$ is a base distribution over the parameter $\theta_n$ of the observation distribution $F(y_n|\theta_n)$. Since the Dirichlet process induces clustering of the parameters $\theta$ into $K$ distinct values, we can equivalently express this model in terms of a distribution over cluster assignments, $x_n \in \{1,\ldots,C\}$. The distribution over $x$ is given by the Chinese restaurant process \citep{aldous85}:
\begin{align}
P(x_n=c|x_{1:n-1}) \propto
  \begin{cases}
   t_c & \text{if } k \leq C_{+} \\
   \alpha       & \text{if } c = C_{+} + 1,
  \end{cases}
\end{align}
where $t_c$ is the number of data points prior to $n$ assigned to cluster $c$ and $C_{+}$ is the number of clusters for which $t_c > 0$.

\subsubsection{Synthetic data}

We first demonstrate our approach on synthetic datasets drawn from various mixtures of bivariate Gaussians (see Table \ref{dpmm_results}). The model parameters for each simulated dataset were chosen to create a spectrum of increasingly overlapping clusters. In particular, we constructed models out of the following building blocks:
\begin{align}
&\mu_1=\bigl(\begin{smallmatrix} 0.0,&0.0 \end{smallmatrix} \bigr), \quad \quad \mu_2=\bigl(\begin{smallmatrix} 0.5,&0.5 \end{smallmatrix} \bigr) \nonumber \\
&\Sigma_1=\bigl(\begin{smallmatrix} 0.25,&0.0\\ 0.0,&0.25 \end{smallmatrix} \bigr), \quad \quad \Sigma_2=\bigl(\begin{smallmatrix} 0.5,&0.0\\ 0.0,&0.5 \end{smallmatrix} \bigr). \nonumber
\end{align}
For the DPMM, we used a Normal likelihood with a Normal-Inverse-Gamma prior on the component parameters:
\begin{align}
y_{nd}|x_n=k \sim \mathcal{N}(m_{kd},\sigma_{kd}^2), \quad \quad m_{kd} \sim \mathcal{N}(0,\sigma_{kd}^2/\tau), \quad \quad \sigma_{kd}^2 \sim \text{IG}(a,b),
\end{align}
where $d \in \{1,2 \}$ indexes observation dimensions and $\text{IG}(a,b)$ denotes the Inverse Gamma distribution with shape $a$ and scale $b$. We used the following hyperparameter values: $\tau=25, a=1, b = 1, \alpha = 0.5$. 

\begin{table*}
\centering
\begin{tabular}{|l|c|c|c|}
Dataset & Particle Filtering ($K=20$) & DPVI ($K=1$) & DPVI ($K=20$) \\
\hline
D1: $[\mu_1,4\mu_2,8\mu_2],\Sigma_1$    & 0.97$\pm$0.03   & 0.93$\pm$0.05   & \textbf{0.99$\pm$0.02}\\
D2: $[\mu_1,4\mu_2,8\mu_2],\Sigma_2$   & 0.89$\pm$0.05   & 0.86$\pm$0.07   & \textbf{0.90$\pm$0.03}\\
D3: $[\mu_1,2\mu_2,4\mu_2],\Sigma_1$   & 0.58$\pm$0.12   & 0.51$\pm$0.03   & \textbf{0.74$\pm$0.16}\\
D4: $[\mu_1,2\mu_2,4\mu_2],\Sigma_2$   & 0.50$\pm$0.06   & 0.46$\pm$0.05   & \textbf{0.55$\pm$0.07}\\
D5: $[\mu_1,\mu_2,2\mu_2],\Sigma_1$   & 0.05$\pm$0.05   & 0.014$\pm$0.02   & \textbf{0.14$\pm$0.10}\\
D6: $[\mu_1,\mu_2,2\mu_2],\Sigma_2$  & 0.15$\pm$0.08   & 0.11$\pm$0.06   & \textbf{0.19$\pm$0.07}
\end{tabular}
\caption{\textbf{Clustering accuracy (V-Measure) for DPMM.} Each dataset consisted of 200 points drawn from a mixture of 3 Gaussians. For each dataset, we repeated the experiment 150 times by iterating through random seeds. The left column shows the ground truth mean for each cluster and the covariance matrix (shared across clusters).}
\label{dpmm_results}
\end{table*} 

\begin{figure*}
\hspace{10mm}  Ground Truth \hspace{15mm} Particle Filter \hspace{20mm} DPVI\\
\centering
(D1)\includegraphics[scale=0.2]{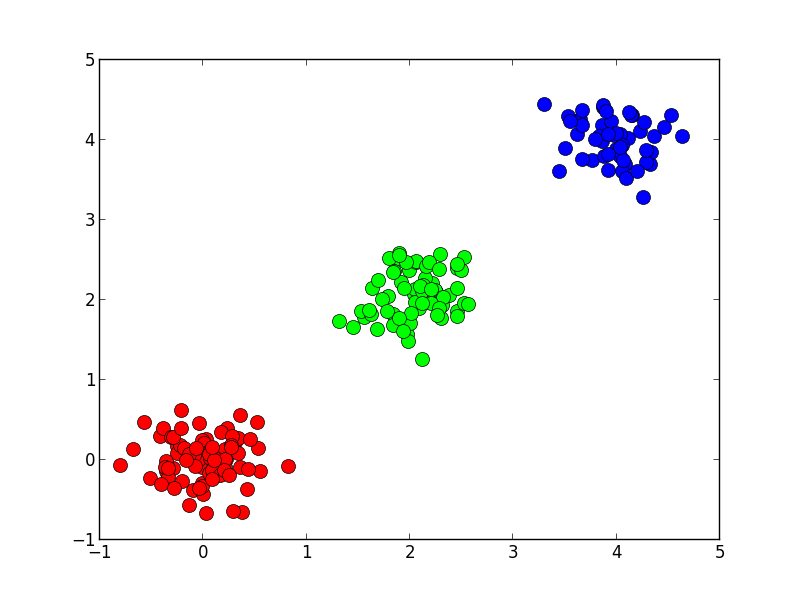}
\includegraphics[scale=0.2]{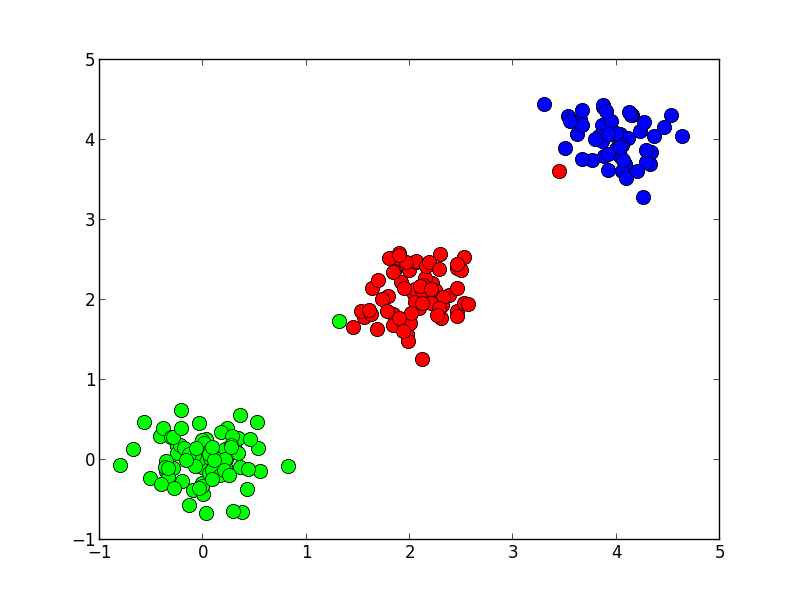}
\includegraphics[scale=0.2]{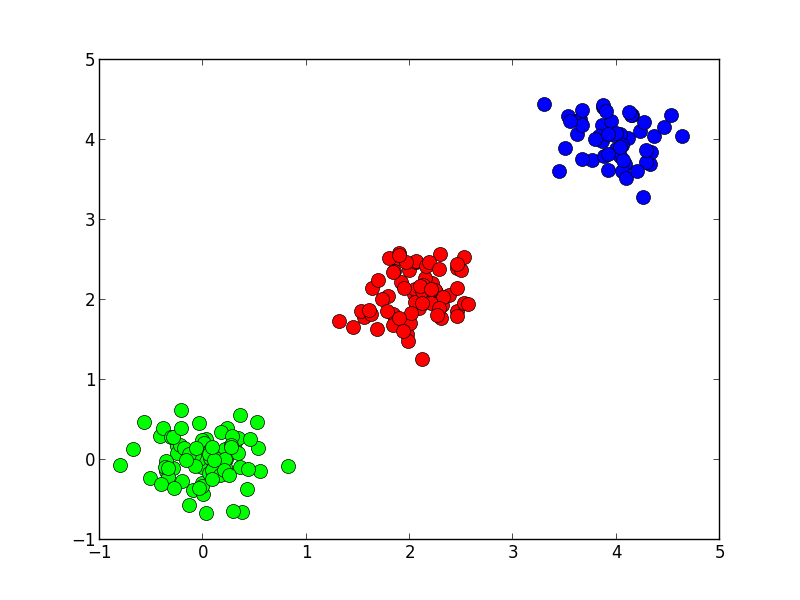}

(D2)\includegraphics[scale=0.20]{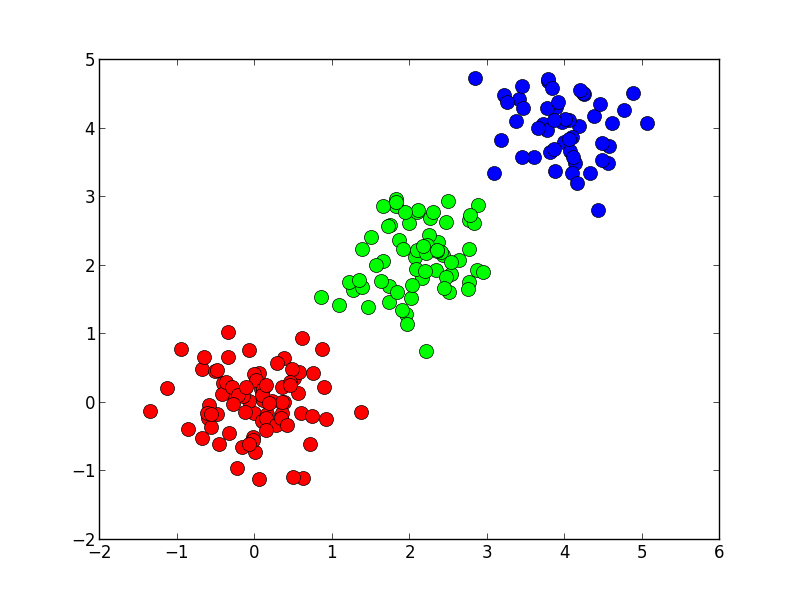}
\includegraphics[scale=0.2]{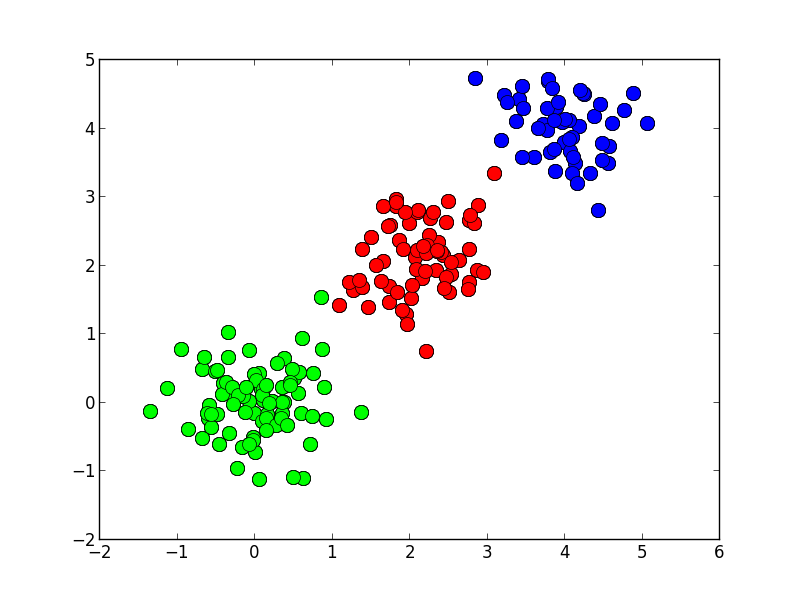}
\includegraphics[scale=0.2]{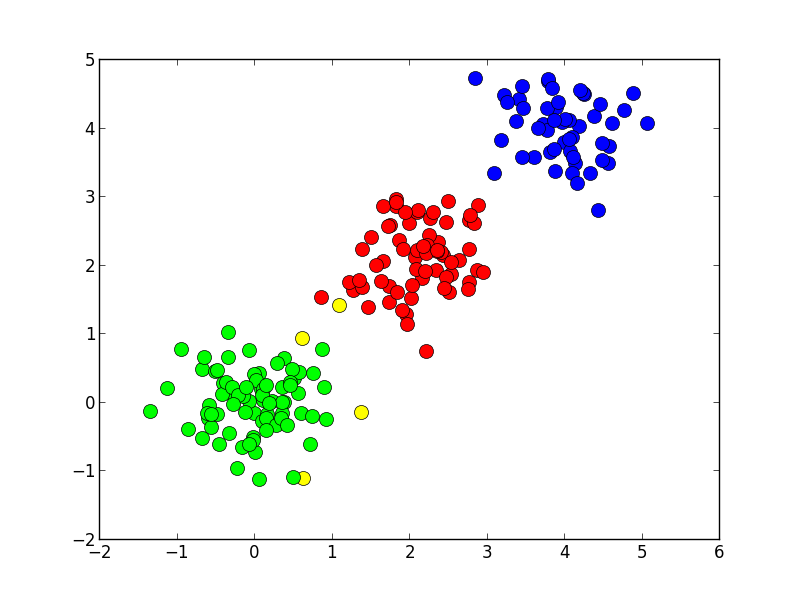}

(D3)\includegraphics[scale=0.2]{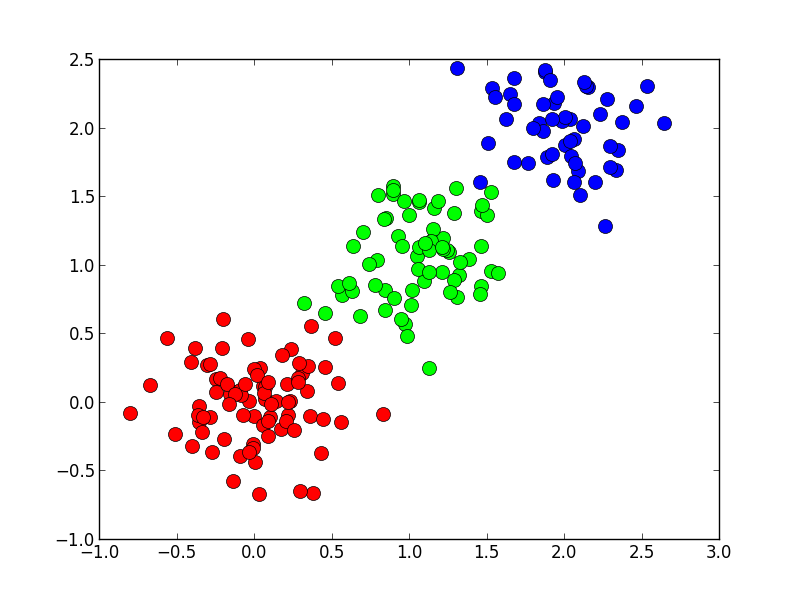}
\includegraphics[scale=0.2]{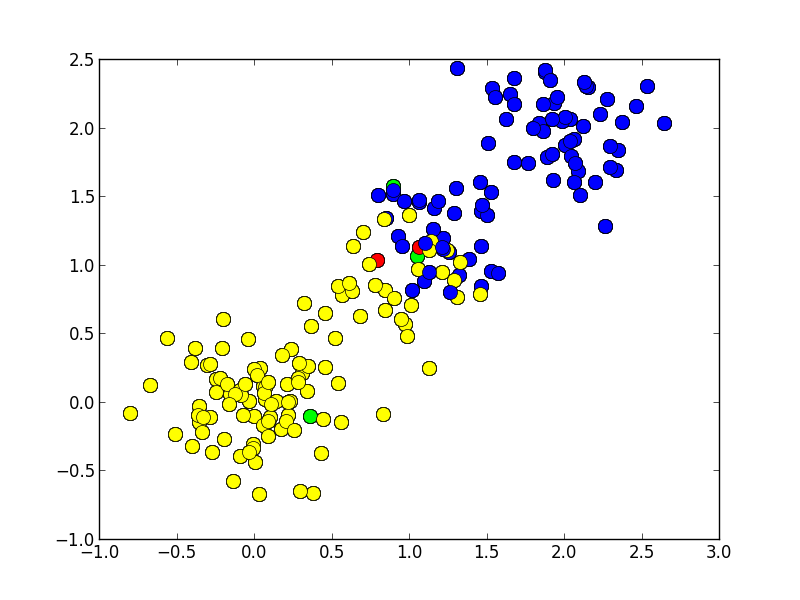}
\includegraphics[scale=0.2]{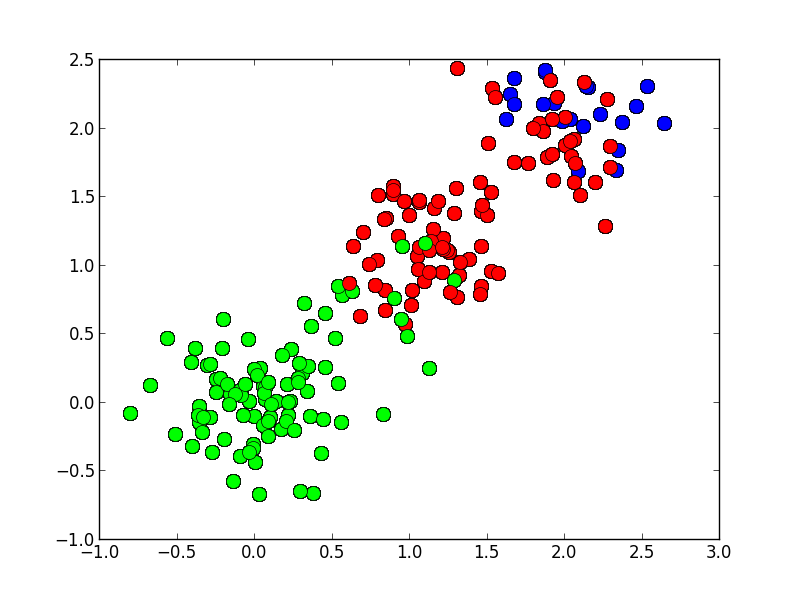}

(D4)\includegraphics[scale=0.2]{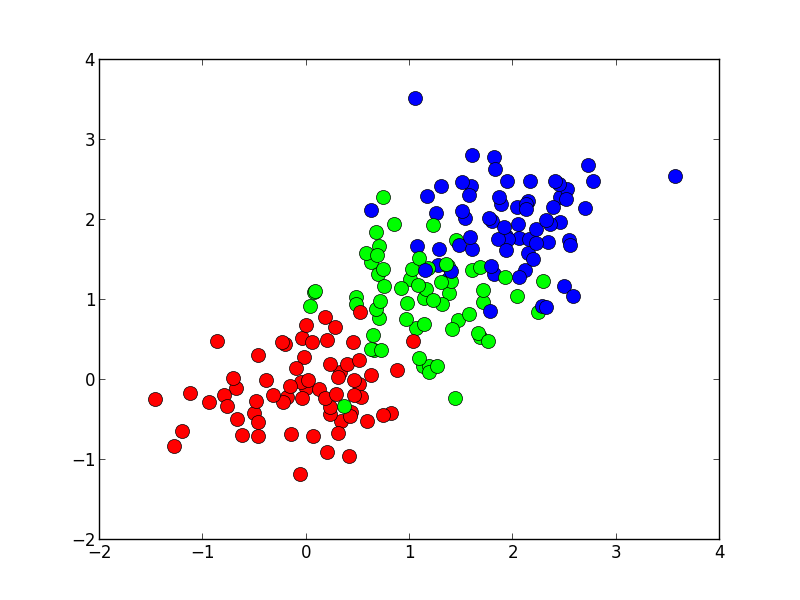}
\includegraphics[scale=0.2]{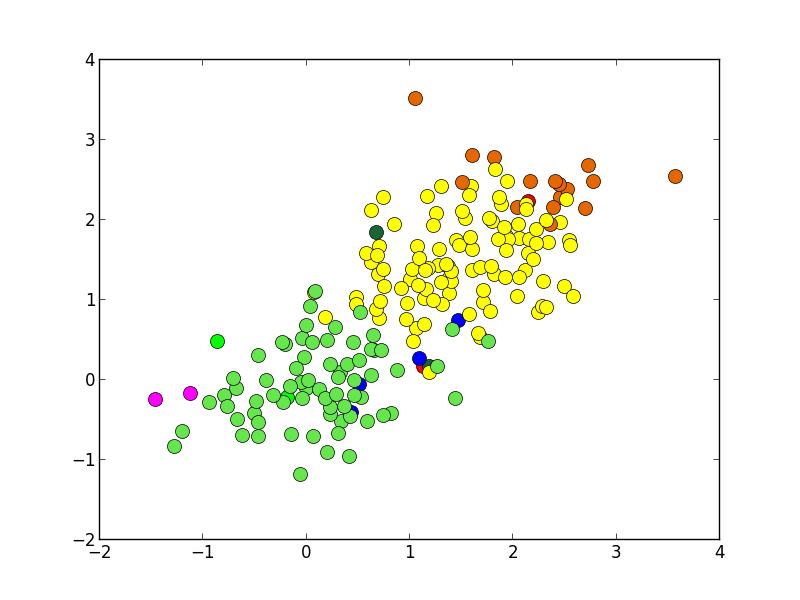}
\includegraphics[scale=0.2]{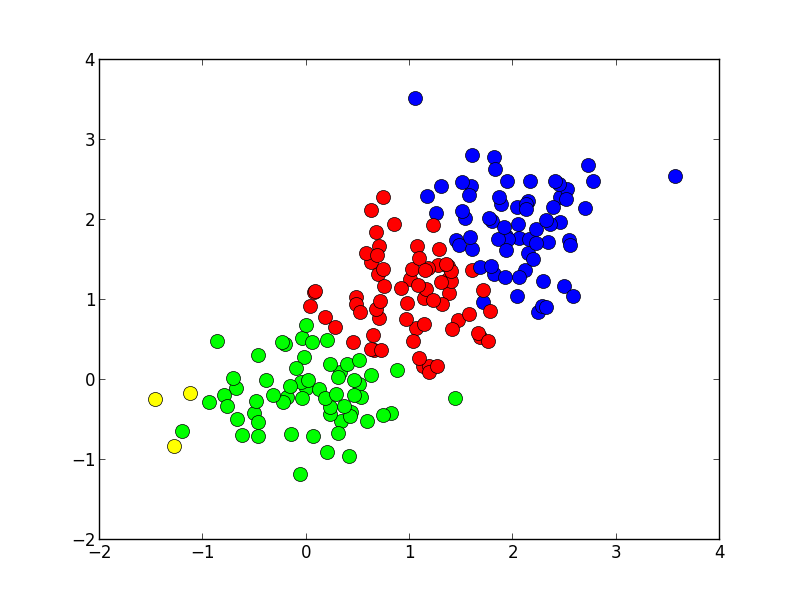}

(D5)\includegraphics[scale=0.2]{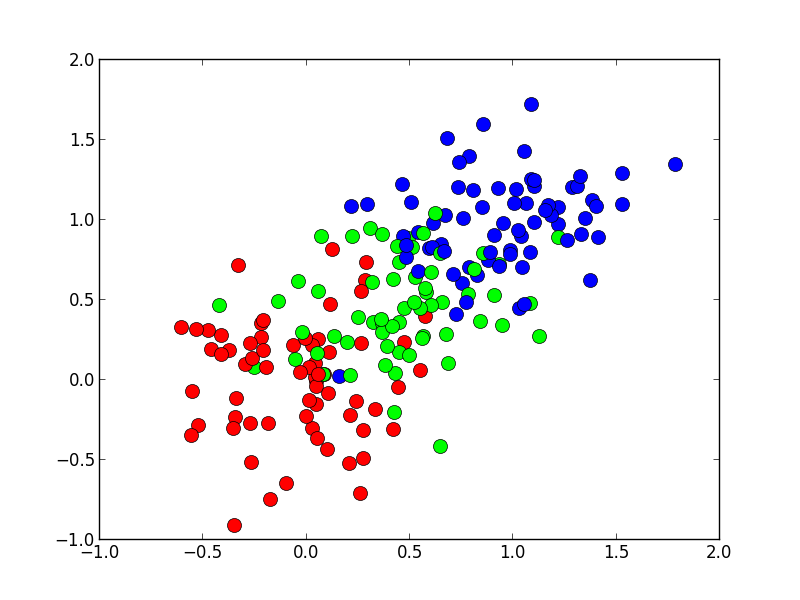}
\includegraphics[scale=0.2]{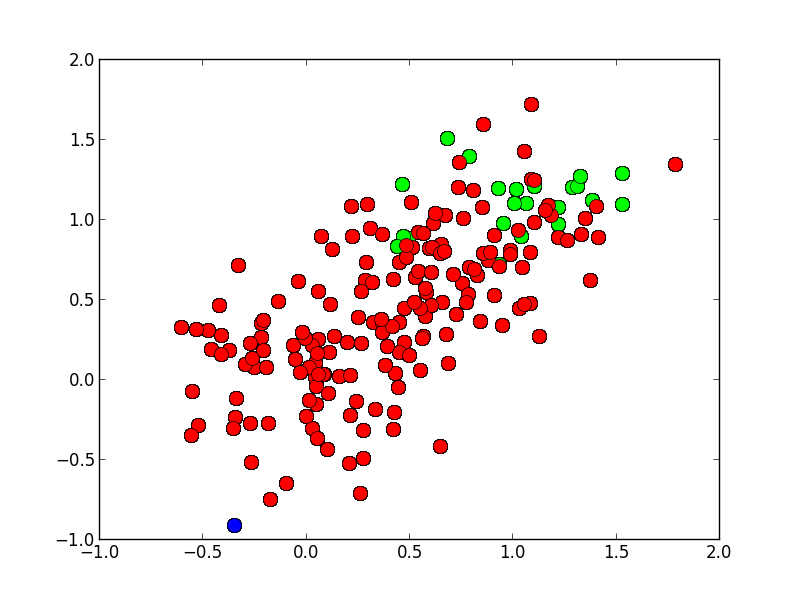}
\includegraphics[scale=0.2]{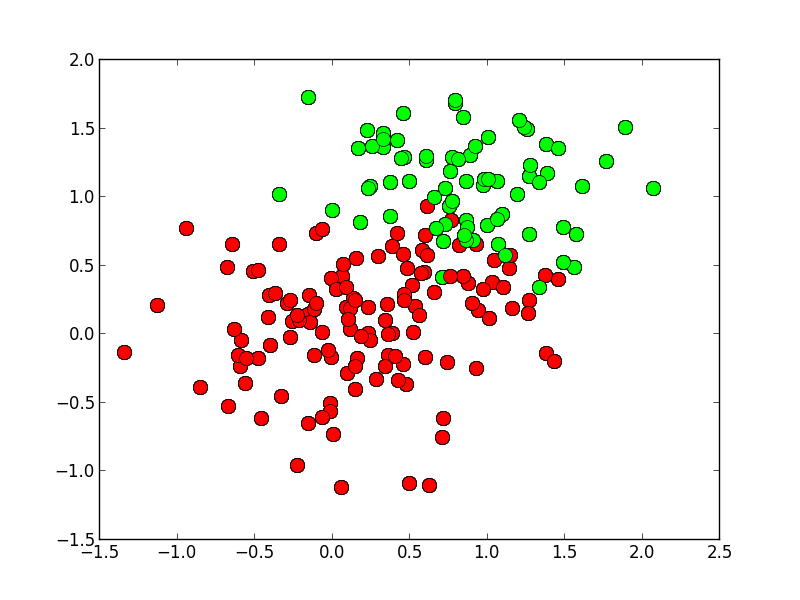}

(D6)\includegraphics[scale=0.2]{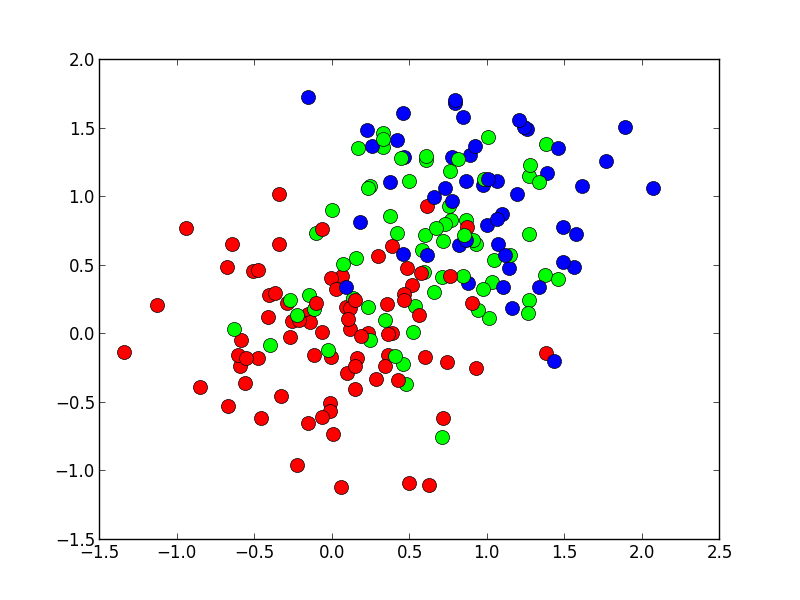}
\includegraphics[scale=0.2]{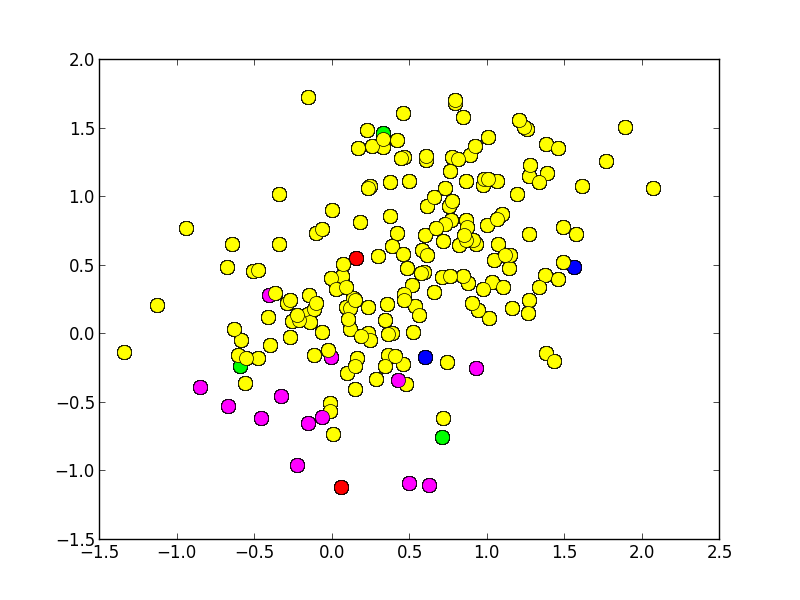}
\includegraphics[scale=0.2]{figures/supplemental/6/maxf.png}

\caption{\textbf{DPMM clustering of synthetic datasets}. We treat DPMM as a filtering problem, analyzing one randomly chosen data point at a time. Colors indicate cluster assignments. Each row corresponds to one synthetic dataset; refer to Table 1 for corresponding quantitative results. Column 1: Ground truth; Column 2: particle filtering; Column 3: DPVI. The DPVI filter scales similarly to the particle filter but does not underfit as severely.}
\label{dpmm_synthetic}
\end{figure*}

Clustering accuracy was measured quantitatively using V-measure \citep{rosenberg07}. Figure \ref{dpmm_synthetic} graphically demonstrates the discovery of latent clusters for both DPVI as well as particle filtering. As shown in Table \ref{dpmm_results}, we observe only marginal improvements when the means are farthest from each other and variances are small, as these parameters leads to well-separated clusters in the training set. However, the relative accuracy of DPVI increases considerably when the clusters are overlapping, either due to the fact that the means are close to each other or the variances are high. 

An interesting special case is when $K=1$. In this case, DPVI is equivalent to the greedy algorithm proposed by \citet{daume07} and later extended by \citet{wang11}. In fact, this algorithm was independently proposed in cognitive psychology by \citet{anderson91}. As shown in Table \ref{dpmm_results}, DPVI with 20 particles outperforms the greedy algorithm, as well as particle filtering with 20 particles.

\subsubsection{Spike sorting}

Spike sorting is an important problem in experimental neuroscience settings where researchers collect large amounts of electrophysiological data from multi-channel tetrodes. The goal is to extract from noisy spike recordings attributes such as the number of neurons, and cluster spikes belonging to the same neuron. This problem naturally motivates the use of DPMM, since the number of neurons recorded by a single tetrode is unknown. Previously, \citet{wood2008nonparametric} applied the DPMM to spike sorting using particle filtering and Gibbs sampling. Here we show that DPVI can outperform particle filtering, achieving high accuracy even with a small number of particles.

We used data collected from a multiunit recording from a human epileptic patient \citep{quiroga2004unsupervised}. The raw spike recordings were preprocessed following the procedure proposed by \citet{quiroga2004unsupervised}, though we note that our inference algorithm is agnostic to the choice of preprocessing. The original data consist of an input vector with $D=10$ dimensions and 9196 data points. Following \citet{wood2008nonparametric}, we used a Normal likelihood with a Normal-Inverse-Wishart prior on the component parameters:
\begin{align}
\mathbf{y}_n|x_n=k \sim \mathcal{N}(\mathbf{m}_k,\Lambda_k), \quad \quad \mathbf{m}_{k} \sim \mathcal{N}(0,\Lambda_k/\tau), \quad \quad \Lambda_k \sim \text{IW}(\Lambda_0,\nu),
\end{align}
where $\text{IW}(\Lambda_0,\nu)$ denotes the Inverse Wishart distribution with degrees of freedom $\nu$ and scale matrix $\Lambda_0$. We used the following hyperparameter values: $\nu=D+1, \Lambda_0 = \mathbf{I}, \tau=0.01, \alpha = 0.1$. 

\begin{figure}[t]
\centering
(A)\includegraphics[width=7.6cm, height=4cm]{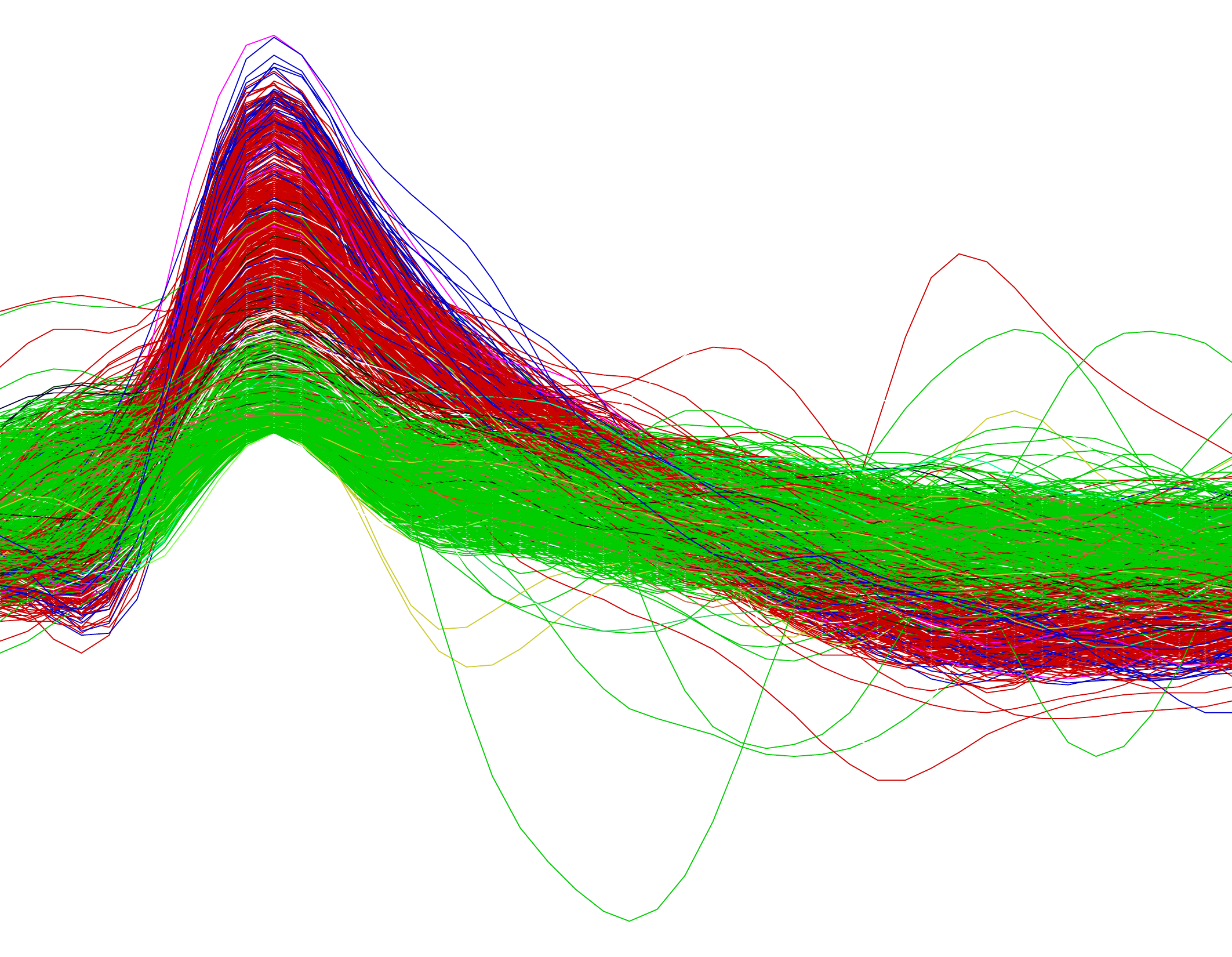}
(B)\includegraphics[width=7.6cm, height=4cm]{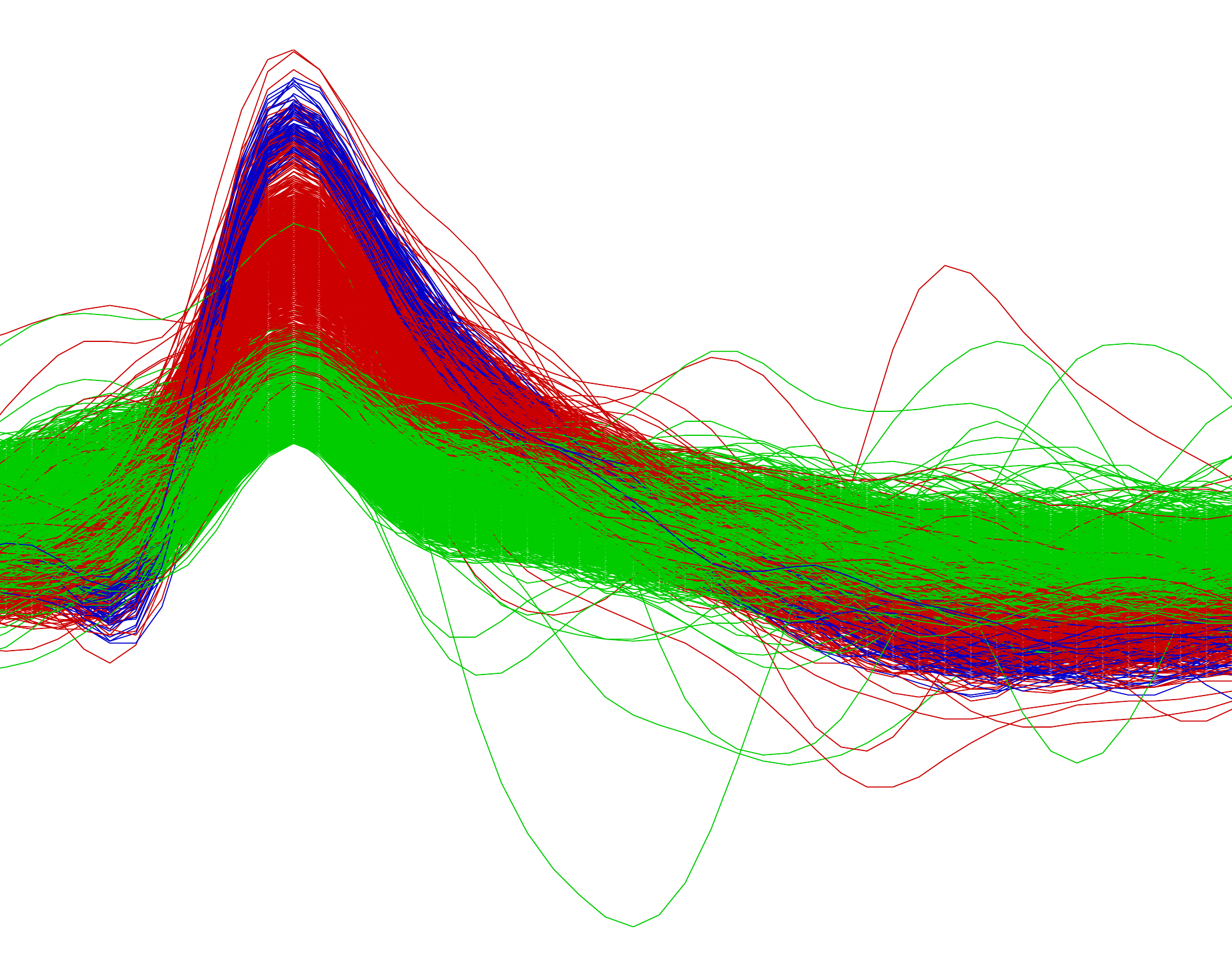}
\caption{\textbf{Spike Sorting using the DPMM}. Each line is an individual spike waveform, colored according to the inferred cluster. (\emph{A}) Result using particle filtering with 100 particles and stratified resampling as reported in \citet{wood2008nonparametric}. (\emph{B}) Result using DPVI. The same model parameters were used for both particle filtering and DPVI.}
\label{spikesort_main}
\end{figure}

We compared our algorithm to the current best particle filtering baseline, which uses stratified resampling \citep{wood2008nonparametric, fearnhead2004particle}. The same model parameters were used for all comparisons. Qualitative results, shown in Figure \ref{spikesort_main}, demonstrate that DPVI is better able to separate the spike waveforms into distinct clusters, despite running DPVI with 10 particles and particle filtering with 100 particles. We also provide quantitative results by calculating the held-out log-likelihood on an independent test set of spike waveforms. The quantitative results (summarized in Table \ref{spikesort_results}) demonstrate that even with only 10 particles DPVI can outperform particle filtering with $1000$ particles. 

\begin{table}
\begin{center}
\begin{tabular}{l|l}
{\bf Method}  &{\bf Held-out log-likelihood} \\
\hline
    DPVI ($K=10$)
    & -3.2474$\times10^5$ ($\hat{C} = 3$)
    \\
    DPVI ($K=100$)
    & \textbf{-1.3888}$\mathbf{\times10^5}$ ($\mathbf{\hat{C} = 3}$)
    \\
    Particle Filtering (Stratified) ($K=10$) 
    & -1.4771$\pm0.21 \times 10^6$ ($\hat{C} = 37$)
    \\
    Particle Filtering (Stratified) ($K=100$) 
    & -5.6757$\pm1.14 \times 10^5$ ($\hat{C} = 13$)
    \\
    Particle Filtering (Stratified) ($K=1000$)
    & -3.2965$\times 10^5$ ($\hat{C} = 5$)
\end{tabular}
\end{center}
\caption{Spike sorting held-out log-likelihood scores for 200 test points. The best performance is achieved by DPVI with 100 particles. Shown in parentheses is the \emph{maximum a posteriori} number of clusters, $\hat{C}$.}
\label{spikesort_results}
\end{table}

\begin{table}
\begin{center}
(A)\begin{tabular}{l|l|c}
{\bf Number of particles} &{\bf DPVI} &{\bf Particle Filtering} \\
\hline
    $K=10$ & 15.20s & 14.71s \\
    $K=50$ & 153.75s & 184.17s \\
	$K=100$ & 567.84s & 699.43s
\end{tabular}
(B)\begin{tabular}{l|l|c} \\
{\bf Number of particles} &{\bf DPVI} &{\bf Particle Filtering} \\
\hline
    $K=10$ & 36.20s & 124s \\
    $K=50$ & 144.6s & 334.2s \\
	$K=100$ & 313.8s & 454.2s
\end{tabular}
\end{center}
\caption{\textbf{Run time comparison for DPMM.} (\emph{A}) Results using synthetic DPMM dataset from Table 1 and (\emph{B}) highlights results obtained by using the spike sorting dataset. In both cases, the run time of DPVI is slightly better than particle filtering.}
\label{runtime_dpmm}
\end{table}

\subsection{Infinite HMM}

An iHMM generates data from the following process \citep{teh06}:
\begin{align}
&G_0 \sim \mbox{DP}(\gamma,H), \quad \quad G_k|G_0 \sim \mbox{DP}(\alpha,G_0), \nonumber \\
&x_n|x_{n-1} \sim G_{x_{n-1}}, \quad \quad \theta_k \sim H, \quad \quad y_n|x_n \sim F(\theta_{x_n}). \nonumber
\end{align}

Like the DPMM, the iHMM induces a sequence of cluster assignments. The distribution over cluster assignments is given by the Chinese restaurant franchise \citep{teh06}. Letting $t_{jc}$ denote the number of times cluster $j$ transitioned to cluster $c$, $x_n$ is assigned to cluster $c$ with probability proportional to $t_{x_{n-1}c}$, or to a cluster never visited from $x_{n-1}$ ($t_{x_{n-1}c}=0$) with probability proportional to $\alpha$. If an unvisited cluster is selected, $x_n$ is assigned to cluster $c$ with probability proportional to $\sum_j t_{jc}$, or to a new cluster (i.e., one never visited from any state, $\sum_j t_{jc}=0$) with probability proportional to $\gamma$.

\begin{figure}
\centering
(A)\includegraphics[scale=0.5]{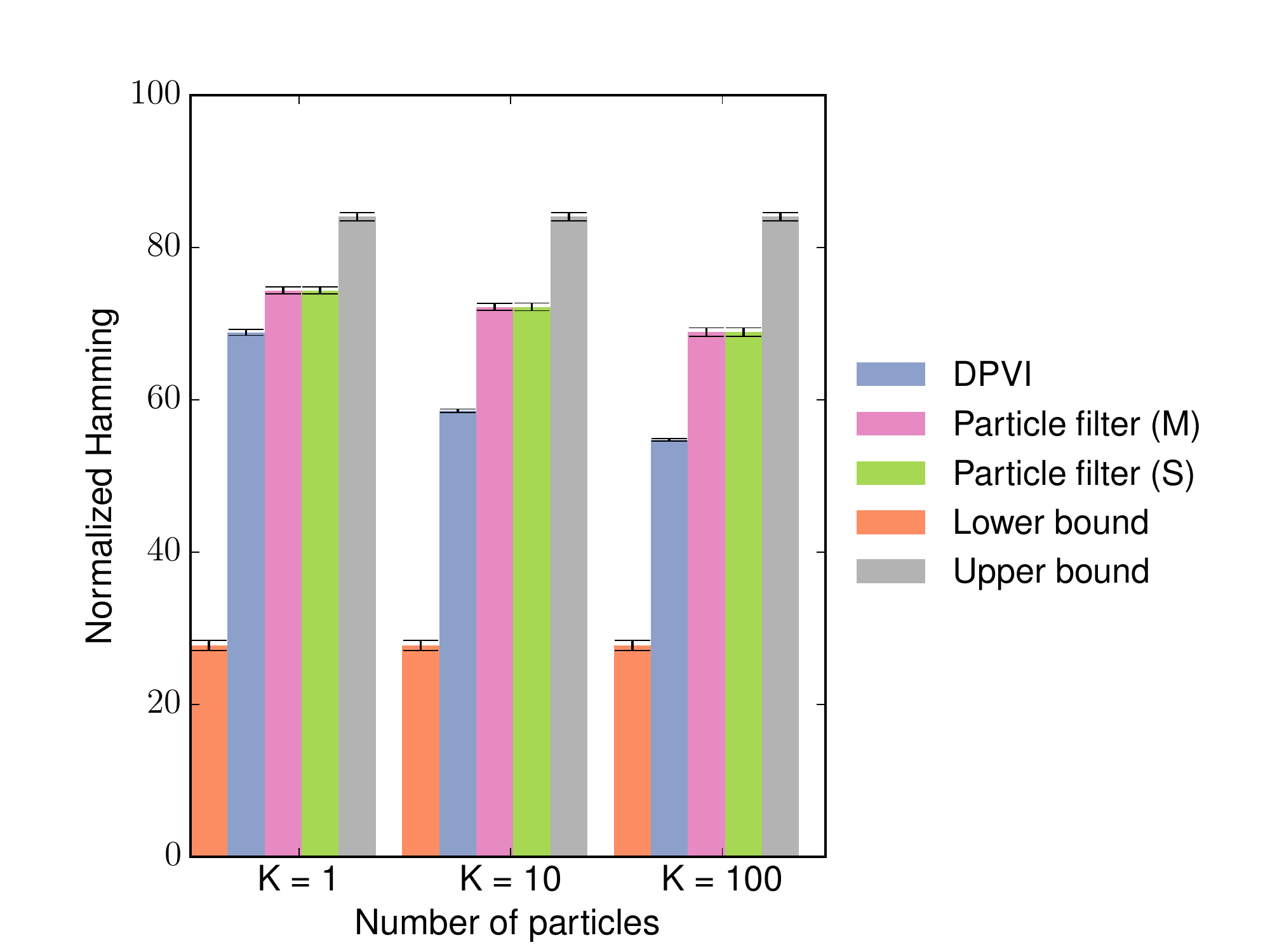}
(B)\includegraphics[scale=0.5]{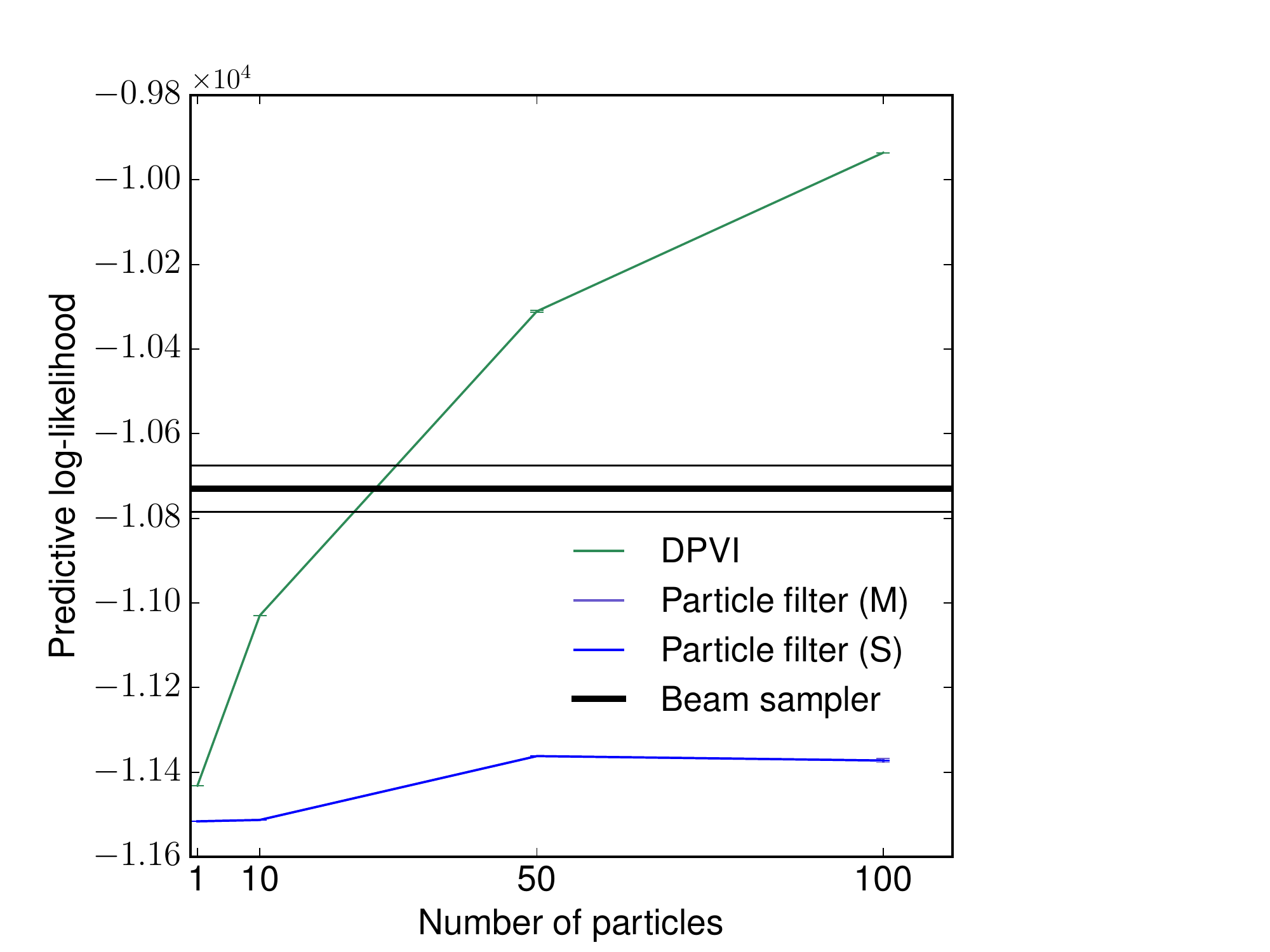}
\caption{\textbf{Infinite HMM results}. (\emph{A}) Results on 500 synthetic data points generated from an HMM with 10 hidden states. Error is the Hamming distance between the true hidden sequence and the sampled sequence, averaged over 50 datasets. M: multinomial resampling; S: stratified resampling. Lower bound is the expected Hamming distance between data-generating distribution and ground truth. Upper bound is the expected Hamming distance between uniform distribution and ground truth. (\emph{B}) Predictive log-likelihood for the ``Alice in Wonderland'' dataset. Particle filtering (M) and (S) overlap in the figure. The error bars in both parts show standard error.}
\label{fig:ihmm_synth}
\end{figure}

\subsubsection{Synthetic data}

We generated 50 sequences with length 500 from 50 different HMMs, each with 10 hidden and 5 observed states. For the rows of the transition and initial probability matrices of the HMMs we used a symmetric Dirichlet prior with concentration parameter 0.1; for the emission probability matrix, we used a symmetric Dirichlet prior with concentration parameter 10.

Figure~\ref{fig:ihmm_synth}A illustrates the performance of DPVI and particle filtering (with multinomial and stratified resampling) for varying numbers of particles ($K=1, 10, 100$). Performance error was quantified by computing the Hamming distance between the true hidden sequence and the sampled sequence. The Munkres algorithm was used to maximize the overlap between the two sequences. The results show that DPVI outperforms particle filtering in all three cases.

When the data consist of long sequences, resampling at every step will produce degeneracy in particle filtering; this tends to result in a smaller number of clusters relative to DPVI. The superior accuracy of DPVI suggests that a larger number of clusters is necessary to capture the latent structure of the data. Not surprisingly, this leads to longer run times (Table \ref{runtime_ihmm}), but it is important to note that particle filtering and DPVI have comparable per-cluster time complexity. 

\begin{table}
\begin{center}
(A)\begin{tabular}{l|l|c}
{\bf Number of particles} &{\bf DPVI} &{\bf Particle Filtering} \\
\hline
   $K=1$ & 1.28 & 1.14s \\
   $K=10$ & 3.56s & 1.92s \\
	$K=100$ & 204.42s & 31.99s
\end{tabular}
(B)\begin{tabular}{l|l|c}\\
{\bf Number of particles} &{\bf DPVI} &{\bf Particle Filtering} \\
\hline
   $K=1$ & 4.73s & 1.64s \\
   $K=10$ & 41.62s & 28.08s \\
	$K=100$ & 1685s & 211.66s
\end{tabular}
\end{center}
\caption{\textbf{Run time comparison for iHMM.} (\emph{A}) Results using the synthetic iHMM dataset from Figure 5A and (\emph{B}) results using the ``Alice in Wonderland'' dataset.}
\label{runtime_ihmm}
\end{table}

\subsubsection{Text analysis}

We next analyzed a real-world dataset, text taken from the beginning of ``Alice in Wonderland'', with 31 observation symbols (letters). We used the first 1000 characters for training, and the subsequent 4000 characters for test. Performance was measured by calculating the predictive log-likelihood. We fixed the hyperparameters  $\alpha$ and $\gamma$ to 1 for both DPVI and the particle filtering. 
 
We ran one pass of DPVI (filtering) and particle filtering over the training sequence. We then sampled 50 datasets from the distribution over the sequences. We truncated the number of states and used the learned transition and emission matrices to compute the predictive log-likelihood of the test sequence. To handle the unobserved emissions in the test sequence we used ``add-$\delta$" smoothing with $\delta = 1$. Finally, we averaged over all the 50 datasets.

We also compared DPVI to the beam sampler \citep{van2008beam}, a combination of dynamic programming and slice sampling, which was previously applied to this dataset. For the beam sampler, we followed the setting of~\cite{van2008beam}.  We run the sampler for 10000 iterations and collect a sample of hidden state sequence every 200 iterations. Figure~\ref{fig:ihmm_synth}B shows the predictive log-likelihood for varying numbers of particles. Even with a small number of particles, DPVI can outperform both particle filtering and the beam sampler. 

\subsection{Infinite relational model (IRM)}

The IRM~\citep{kemp2006learning} is a nonparametric model of relational systems. The model simultaneously discovers the clusters of entities and the relationships between the clusters. A key assumption of the model is that each entity belongs to exactly one cluster.

Given a relation $R$ involving $J$ types of entities, the goal is to infer a vector of cluster assignments $x^j$ for all the entities of each type $j=1,\ldots,J$.\footnote{The IRM model can be defined for multiple relations but for simplicity we only describe the single relation case.} Assuming the cluster assignments for each type are independent, the joint density of the relation and the cluster assignment vectors can be written as:
\begin{align}
P(R, x^1, \ldots, x^J) = P(R|x^1,\ldots, x^J) \prod_{j=1}^J P(x^j).
\end{align}
The cluster assignment vectors are drawn from a $\mbox{CRP}(\alpha)$ prior. Given the cluster assignment vectors, the relations are drawn from a Bernoulli distribution with a parameter that depends on the clusters involved in that relation. More formally, let us define a binary relation $R: T^{d_1} \times \dots T^{d_M} \mapsto \{0,1\}$, where $d_m$ is the label of the type occupying position $m$ in the relation. Each relational value is generated according to:
\begin{align}
R(i_1,\ldots,i_M)|x^1,\ldots,x^J \sim \text{Bernoulli}(\eta(x_{i_1}^{d_1},\ldots,x_{i_M}^{d_M})),
\end{align}
where $i_m$ denotes the entity (of type $d_m$) occupying position $m$. Each entry of $\eta$ is drawn from a Beta($\beta,\beta$) distribution. By using a conjugate Beta-Bernoulli model, we can analytically marginalize the parameters $\eta$ \citep[see][]{kemp2006learning}, allowing us to directly compute the likelihood of the relational matrix given the cluster assignments, $P(R|x^1,\ldots,x^J)$.

We compared the performance of DPVI with Gibbs sampling, using predictive log-likelihood on held-out data as a performance metric. Two datasets analyzed in \cite{kemp2006learning}, ``animals'' and ``Alyawarra", were used for this task. The animals dataset \citep{osherson91} is a two type dataset $R:T_1 \times T_2 \to \{0, 1\} $ with animals and features as it types; it contains 50 animals and 85 features. The Alyawarra dataset \citep{denham73} has a ternary relation $R: T_1 \times T_1 \times T_2 \to \{0, 1\}$ where $T_1$ is the set of 104 people and $T_2$ is the set of 25 kinship terms. 

\begin{figure}
\footnotesize
\begin{center}
\begin{tabular}{ll}
\multirow{8}{*}{\includegraphics[trim = 33mm 0mm 15mm 20mm, clip, scale = 0.43]{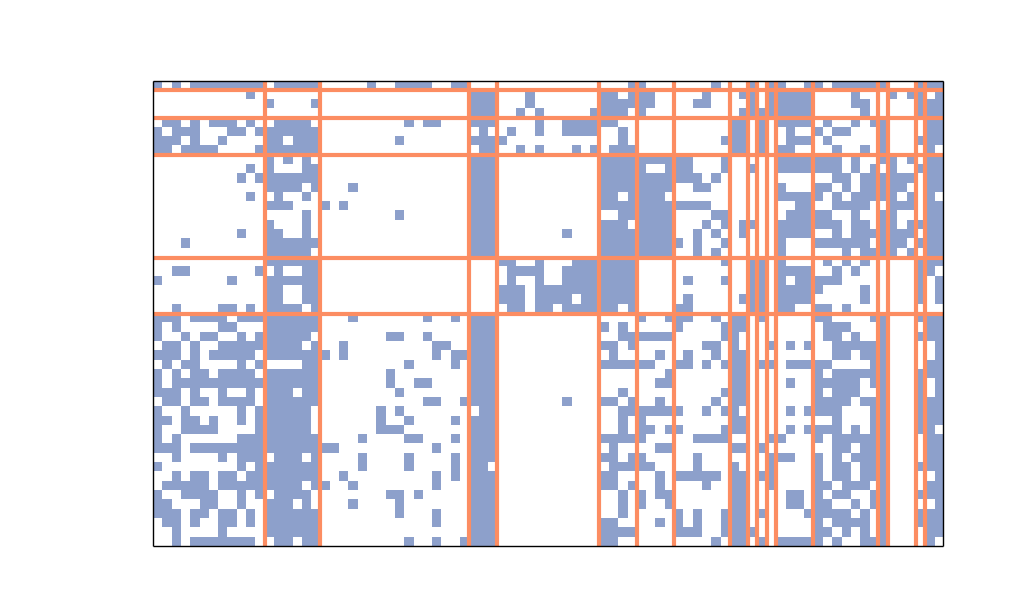}} & \textbf{Sample animal clusters:}\\
& \textbf{A1}: Hippopotamus, Elephant, Rhinoceros\\
& \textbf{A2}: Seal, Walrus, Dolphins, Blue Whale,  \\
& \hspace{6mm} Killer Whale, Humpback Whale\\
& \textbf{A3}: Beaver, Otter, Polar Bear\\
\\
& \textbf{Sample feature clusters:} \\
& \textbf{F1}: Hooves, Long neck, Horns \\
& \textbf{F2}: Inactive, Slow, Bulbous Body, Tough Skin\\
& \textbf{F3}: Lives in Fields, Lives in Plains, Grazer\\
& \textbf{F4}: Walks, Quadrupedal, Ground\\
& \textbf{F5}: Fast, Agility, Active, Tail
\end{tabular}
\vspace{1mm}
\caption{Co-clustering  of animals (rows) and features (columns) after 50 iterations of DPVI with 10 particles.}
\label{fig:IRM_table}
\end{center}
\end{figure}

We removed 20\% of the relations form each dataset and computed the predictive log-likelihood for the held-out data. We ran DPVI with 1, 10 and 20 particles for 100 iterations. Given the weights of the particles, we computed the weighted log-likelihood. We also ran 20 independent runs of the Gibbs sampler for 100 iterations and computed the average predictive log-likelihood. Every iteration scans all the data points in all the types sequentially. We set the hyperparameters $\alpha$ and $\beta$ to 1. Figure \ref{fig:IRM_table} illustrates the co-clustering discovered by DPVI for the animals dataset, demonstrating intuitively reasonable animal and feature clusters. 

\begin{figure}
\vspace{-3mm}
\begin{center}
\begin{tabular}{cc}
\hspace{-5mm}\includegraphics[trim = -2mm 0mm 0mm 0mm, clip, scale = 0.4]{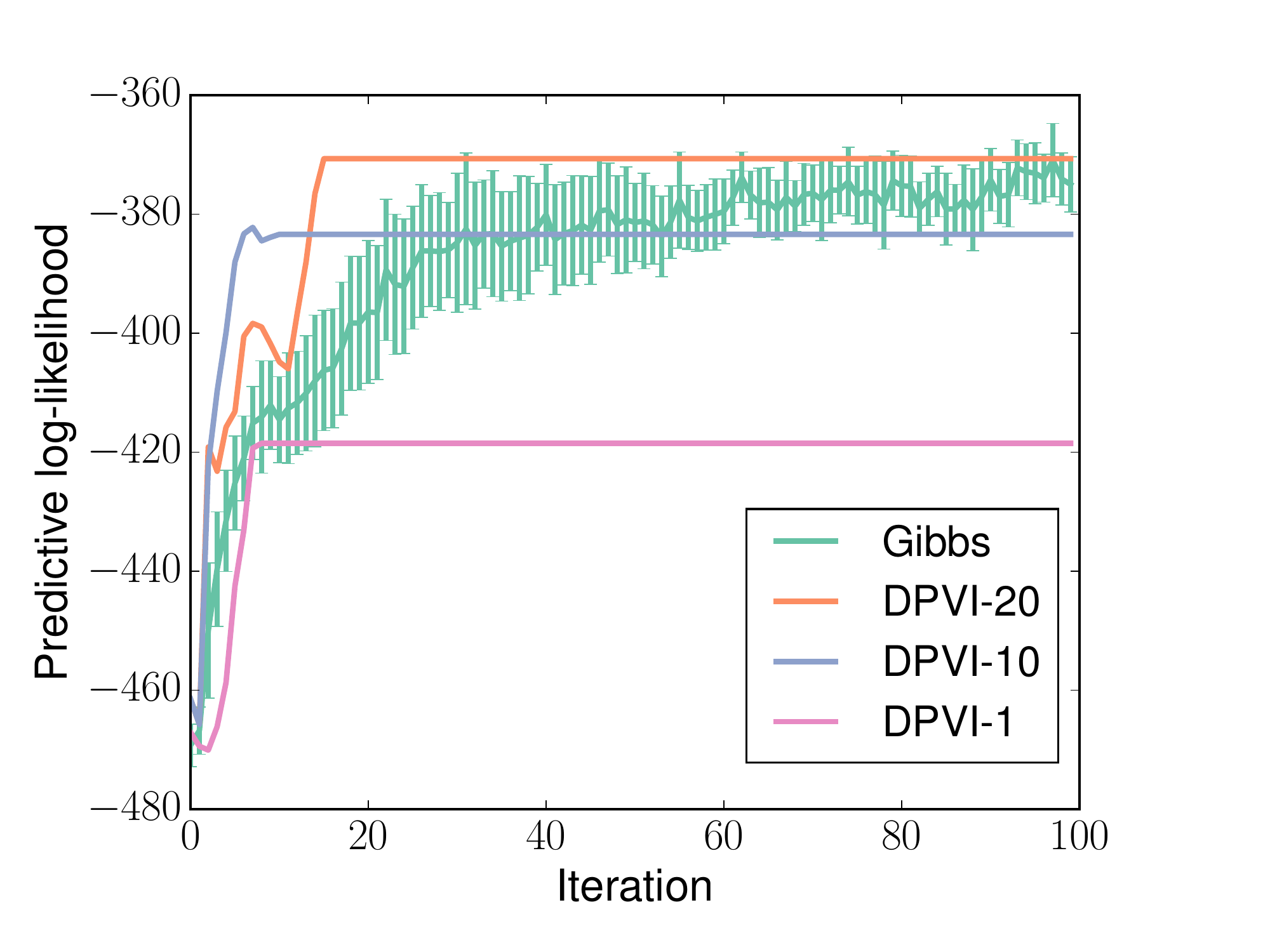} & \hspace{-17mm}
\includegraphics[trim = 19mm 0mm 0mm 0mm, clip, scale = 0.4]{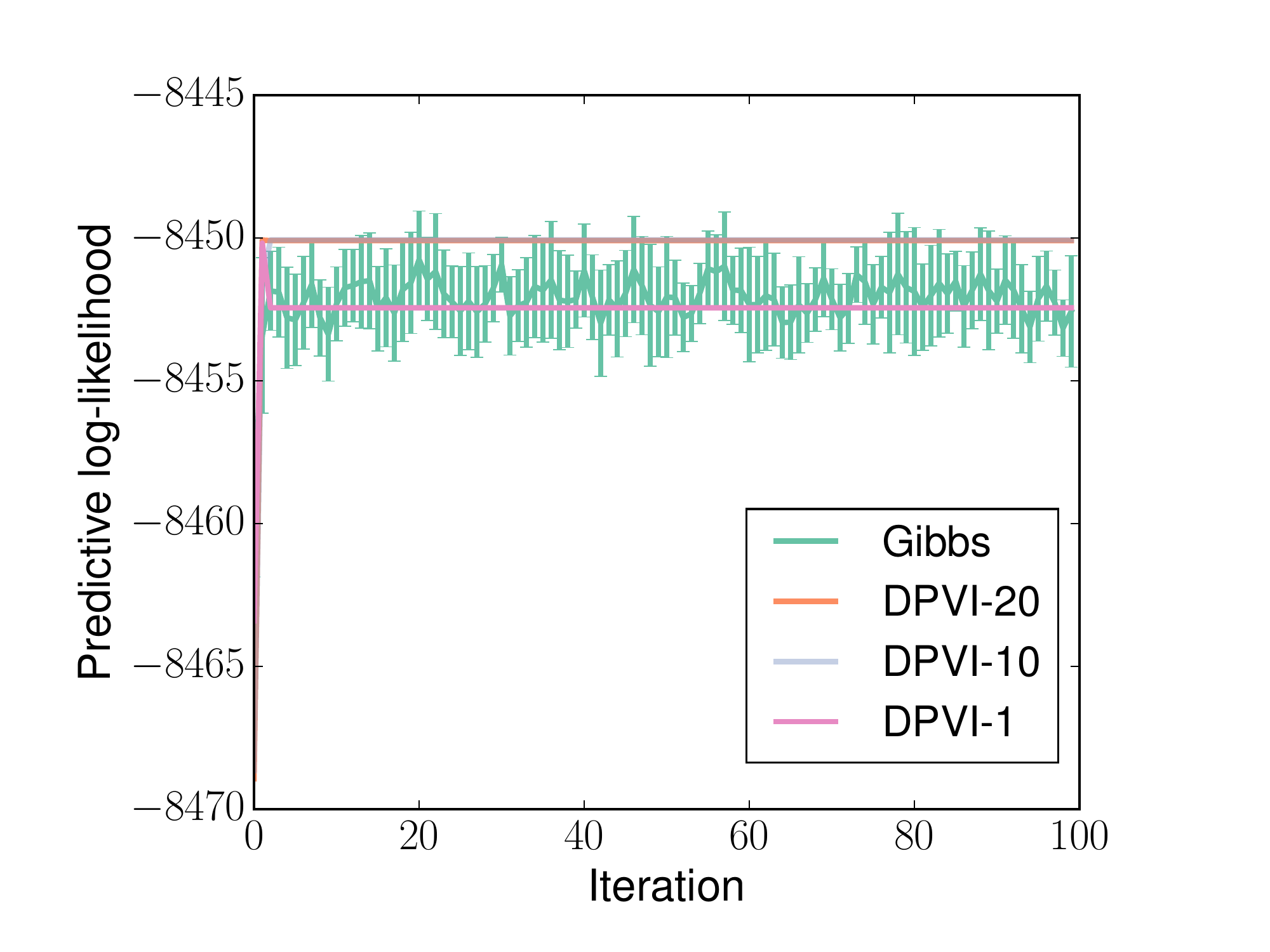}\\
\hspace{-5mm} (A)  &\hspace{-17mm} (B) \\
\end{tabular}
\vspace{-3mm}
\caption{Predictive log-likelihood vs iteration for 
(\emph{A}) Animals and (\emph{B}) Alyawarra datasets. For DPVI the predictive log-likelihood is the weighted average across all the particles. For Gibbs sampling the bold line corresponds to the mean across samples, and the error bars correspond to the standard error.}
\label{fig:IRM_pred}
\end{center}
\vspace{-6mm}
\end{figure}

The results after 100 iterations are presented in Table~\ref{tab:IRM_results}. The best performance is achieved by DPVI with 20 particles. Figure~\ref{fig:IRM_pred} shows the predictive log-likelihood for every iteration of DPVI and Gibbs sampling. For the animals dataset, DPVI with 10 and 20 particles converge in 11 and 18 iterations, respectively. The number of iterations required for convergence in the Alyawarra dataset is just 2 and 3 for 10 and 20 particles, respectively. In terms of computation time per iteration of DPVI versus Gibbs, the only difference for DPVI with one particle and Gibbs is the sorting cost. Hence, for the multiple particle versus multiple runs of Gibbs sampling, the only additional cost is the sorting cost for multiple particles (e.g. 10 or 20). However, this insignificant additional cost is compensated for by a faster convergence rate in our experiments.

\begin{table}
\begin{center}
\begin{tabular}{ l | c c}
& \multicolumn{2}{ c }{\textbf{Predicitive log-likelihood}} \\
{\bf Method}  &{\bf \footnotesize{Animals}} & {\bf \footnotesize{Alyawarra}}\\
\hline
    DPVI ($K=1$)
    & -418.498 & -8.452 $\times10^3$
    \\
    DPVI ($K=10$)
    & -382.543 & \textbf{-8.450 $\times10^3$}
    \\
    DPVI ($K=20$) 
    & \textbf{-370.674} & \textbf{-8.450 $\times10^3$}
    \\
    Gibbs (avg. of $20$ runs) 
    & -374.986 & -8.453 $\times10^3$
    \\
\end{tabular}
\end{center}
\caption{Predictive log-likelihood after 100 iterations of DPVI and Gibbs for the animals and  Alyawarra datasets (with 20 \% held-out). The best performance is achieved by DPVI with 20 particles.}
\label{tab:IRM_results}
\end{table}

\subsection{Ising model}

So far, we have been studying inference in directed graphical models, but DPVI can also be applied to undirected graphical models. We illustrate this using the Ising model for binary vectors $x \in \{-1,+1\}^N$:
\begin{align}
f(x) = \frac{1}{2} x W x^\top + \theta x^\top,
\end{align}
where $W \in \mathbb{R}^{N \times N}$ and $\theta \in \mathbb{R}^N$ are fixed parameters. In particular, we study a square lattice ferromagnet, where $W_{ij} = \beta$ for neighboring nodes (0 otherwise) and $\theta_i=0$ for all nodes. We refer to $\beta$ as the \emph{coupling strength}. This model has two global modes: when all the nodes are set to 1, and when all the nodes are set to 0. As the coupling strength increases, the probability mass becomes increasingly concentrated at the two modes.

\begin{figure}
\centering
\includegraphics[scale=0.3]{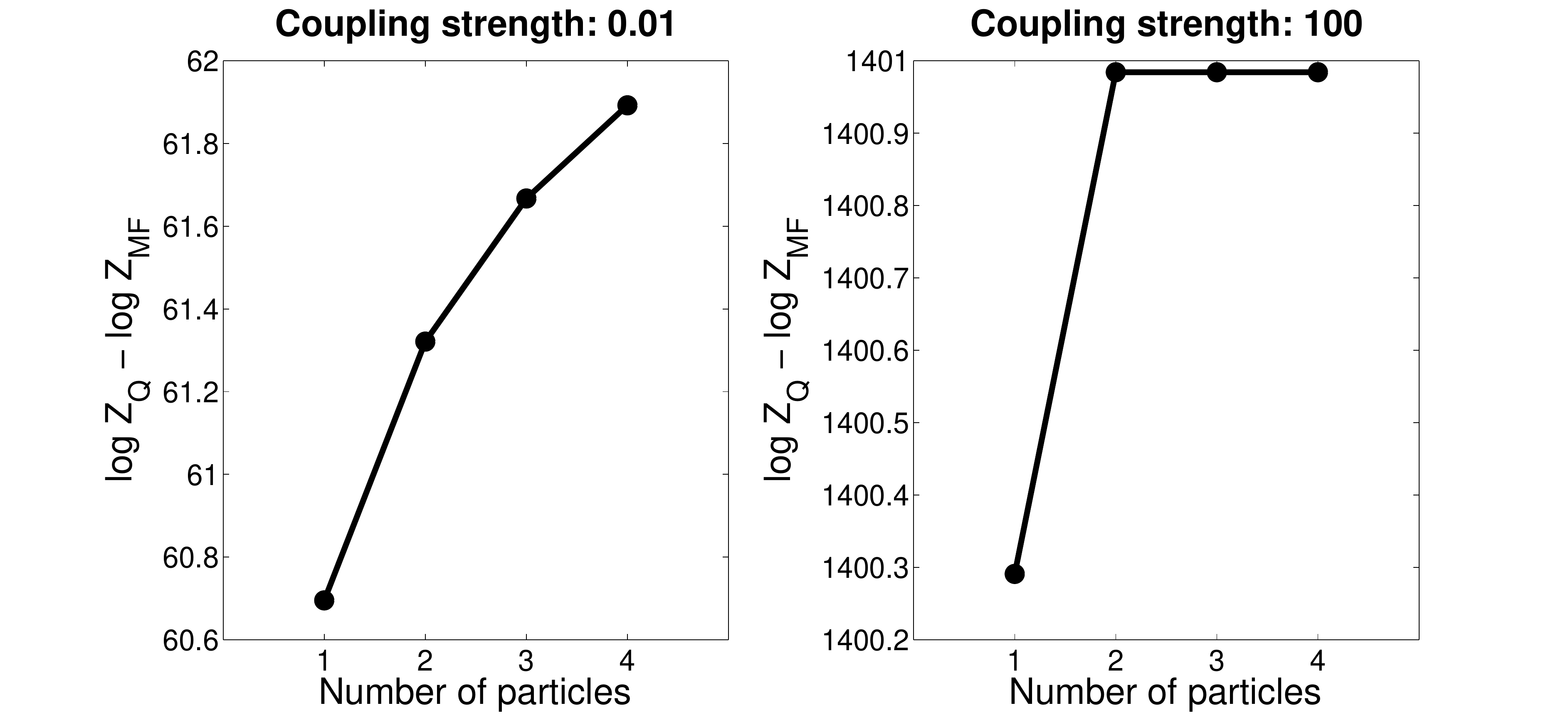}
\caption{\textbf{Ising model results}. Difference between DPVI and mean-field lower bounds on the partition function. Positive values indicates superior DPVI performance. (\emph{A}) Low coupling strength; (\emph{B}) high coupling strength.}
\label{fig:ising}
\end{figure}

We applied DPVI to this model, varying the number of particles and the coupling strength. To quantify performance, we computed the DPVI variational lower bound on the partition function and compared this to the lower bound furnished by the mean-field approximation \citep[see][]{wainwright08}. Figure \ref{fig:ising}A shows the results of this analysis for low coupling strength ($\beta=0.01$) and high coupling strength ($\beta=100$). DPVI consistently achieves a better lower bound than mean-field, even with a single particle, and this advantage is especially conspicuous for high coupling strength. Adding more particles improves the results, but more than 3 particles does not appear to confer any additional improvement for high coupling strength. These results illustrate how DPVI is able to capture multimodal target distributions, where mean-field approximations break down (since they cannot effectively handle multimodality).

\begin{figure}
\centering
\includegraphics[scale=0.8]{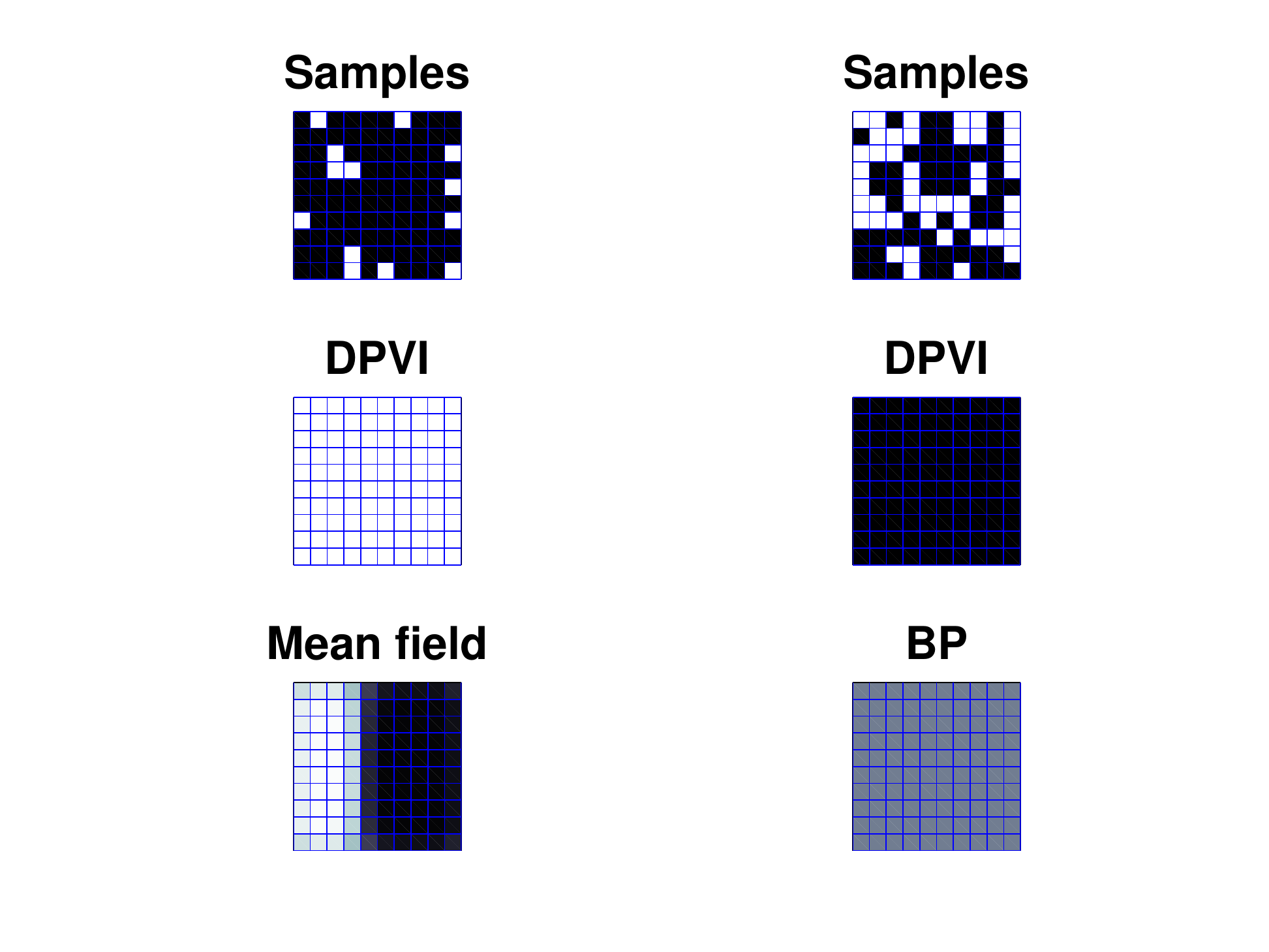}
\caption{\textbf{Ising model simulations}. Examples of posteriors for the ferromagnetic lattice at low coupling strength. (\emph{Top}) Two configurations from a Swendsen-Wang sampler. (\emph{Middle}) Two DPVI particles. (\emph{Bottom left}) Mean-field expected value. (\emph{Bottom right}) Loopy belief propagation expected value.}
\label{fig:ising_approx}
\end{figure}

To illustrate the performance of DPVI further, we compared several posterior approximations for the Ising model in Figure \ref{fig:ising_approx}. In addition to the mean-field approximation, we also compared DPVI with two other standard approximations: the Swendsen-Wang Monte Carlo sampler \citep{swendsen87} and loopy belief propagation \citep{murphy99}. The sampler tended to produce noisy results, whereas mean-field and BP both failed to capture the multimodal structure of the posterior. In contrast, DPVI with two particles perfectly captured the two modes.

\section{Conclusions}

This paper introduced a variational framework for particle approximations of discrete probability distributions. We described a practical algorithm for optimizing the approximation, and showed empirically that it can outperform widely-used Monte Carlo and variational algorithms. The key to the success of this approach is an optimal selection of particles: Rather than generating them randomly (as in Monte Carlo algorithms), we deterministically choose a set of unique particles that optimizes the KL divergence between the approximation and the target distribution. Because we are selecting particles optimally, we can achieve good performance with a smaller number of particles compared to Monte Carlo algorithms, thereby improving computational efficiency. Another advantage of DPVI is that its deterministic nature eliminates the contribution of Monte Carlo variance to estimation error.

A consistent problem vexing sequential Monte Carlo methods like particle filtering is the double-edged sword of resampling: this step is necessary to remove conditionally unlikely particles, but the resulting loss of particle diversity can lead to degeneracy. As we showed in our experiments, tuning an ESS threshold for resampling can improve performance, but requires finding a relatively narrow sweet spot for the threshold. DPVI achieves comparable performance to the best particle filter by using a deterministic strategy for deleting and replacing particles, avoiding finicky tuning parameters. It is also worth noting two other desirable properties of DPVI in this context: (1) the particle set is guaranteed to be diverse because all particles are unique; (2) all the particles have high probability and therefore the propagation of conditionally unlikely particles is avoided, as happens when particle filtering is run without resampling. We believe that this combination of properties is a key to the superior performance of DPVI relative to particle filtering.

An important task for future work is to consider how DPVI can be efficiently applied to models with combinatorial latent structure (such as the factorial HMM), which may have too many assignments to enumerate completely. In this setting, it is desirable to use a proposal distribution to selectively sample certain assignments. An interesting possibility is to use randomly seeded optimization algorithms to generate high probability proposals. Since the proposal mechanism does not play any role in the score function (unlike in particle filtering, where samples have to be reweighted), we are free to choose any deterministic or stochastic proposal mechanism without needing to evaluate its probability density function.

In summary, DPVI harmoniously combines a number of ideas from Monte Carlo and variational methods. The resulting algorithm can achieve performance superior to widely used particle filtering, MCMC and mean-field methods, though more work is needed to evaluate its performance on a wider range of probabilistic models and to compare it to other inference algorithms.

\section*{Acknowledgments}
TDK is generously supported by the Leventhal Fellowship. VKM is supported by the Army Research Office Contract Number 0010363131, Office of Naval Research Award N000141310333, and the DARPA PPAML program. The views and conclusions contained herein are those of the authors and should not be interpreted as necessarily representing the official policies or endorsements, either expressed or implied, of the U.S. Government. The U.S. Government is authorized to reproduce and distribute reprints for Governmental purposes notwithstanding any copyright annotation thereon.

\bibliography{bib}

\end{document}